\documentclass{article} 

\usepackage{iclr2027_conference,times}

\usepackage{microtype}
\usepackage{url}
\usepackage{booktabs}
\usepackage{algorithm}
\usepackage{algpseudocode}
\usepackage{amsmath}
\usepackage{multirow}
\usepackage{graphicx}
\usepackage{subcaption}
\usepackage{tabularx}
\usepackage{listings}
\usepackage[table]{xcolor}
\usepackage{adjustbox}
\usepackage{makecell}
\usepackage{array}
\usepackage{diagbox}
\usepackage{siunitx}

\usepackage{hyperref}

\definecolor{darkblue}{rgb}{0, 0, 0.5}
\hypersetup{
    colorlinks=true,
    citecolor=darkblue,
    linkcolor=darkblue,
    urlcolor=darkblue
}

\definecolor{ocrbg}{RGB}{245,245,245}
\definecolor{basebg}{RGB}{255,243,214}
\definecolor{optbg}{RGB}{222,243,226}
\definecolor{ftbg}{RGB}{222,235,247}
\definecolor{jointbg}{RGB}{236,226,248}
\definecolor{otherbg}{RGB}{240,240,255}
\definecolor{retrieverbg}{RGB}{245,245,245}

\newcommand{\zerodelta}{0.0}
\newcommand{\pos}[1]{\textcolor{green!50!black}{#1}}
\newcommand{\negd}[1]{\textcolor{red!70!black}{#1}}

\newcommand{\deltacell}[1]{\multicolumn{1}{c}{\scriptsize #1}}

\iclrfinalcopy 

\title{Document Optimization for Black-Box \\ Retrieval via Reinforcement Learning}

\author{
Omri Uzan$^{1}$, Ron Polonsky$^{1}$, Douwe Kiela$^{1,2}$, Christopher Potts$^{1}$
\\ \\
$^{1}$ Stanford University,
$^{2}$ ContextualAI \\ \\
uzan@stanford.edu
}

\newcommand{\ptrans}{p_{\textit{trans}}}

\begin{document}

\maketitle

\fancyhead{}
\renewcommand{\headrulewidth}{0pt}

\begin{abstract}
Generative large language models (LLMs) are increasingly used as inference-time components in retrieval pipelines, for tasks such as query rewriting and document reranking. However, these online approaches place costly autoregressive computation directly on the latency-critical retrieval path. We explore an alternative axis: using LLMs to improve documents instead, rewriting them into better representations and shifting computation offline. Yet producing a useful document rewrite is not straightforward: retrieval is inherently discriminative, so an effective rewrite must make a document more similar to relevant queries than competing candidates under the retriever’s notion of similarity. We therefore formulate document transformation as an optimization problem, directly training an LLM or VLM to produce rewrites that improve retrieval. Our approach, DocOpt, uses GRPO with retriever ranking improvements as rewards, requires only black-box access to retrieval ranks, and applies across single-vector, multi-vector, and lexical retrievers. We evaluate zero-shot LLM rewriting and DocOpt on code and visual retrieval tasks, finding that document rewriting can improve retrieval and that optimizing rewrites yields further gains. For example, OpenAI text-embedding-3-small achieves 58.35 nDCG@5 on average with direct retrieval; zero-shot rewriting improves this to 60.83 with GPT-5.4-mini, 63.75 with Claude Haiku 4.5, and 64.23 with Qwen3. DocOpt further improves performance to 67.94, surpassing the 6.5$\times$ more expensive text-embedding-3-large retriever at 66.15.
We release our code at \href{https://github.com/omriuz/doc-opt}{github.com/omriuz/doc-opt}.
\end{abstract}

\section{Introduction}

\begin{figure}[htbp]
    \centering
    \includegraphics[width=\linewidth,height=0.45\linewidth,keepaspectratio]{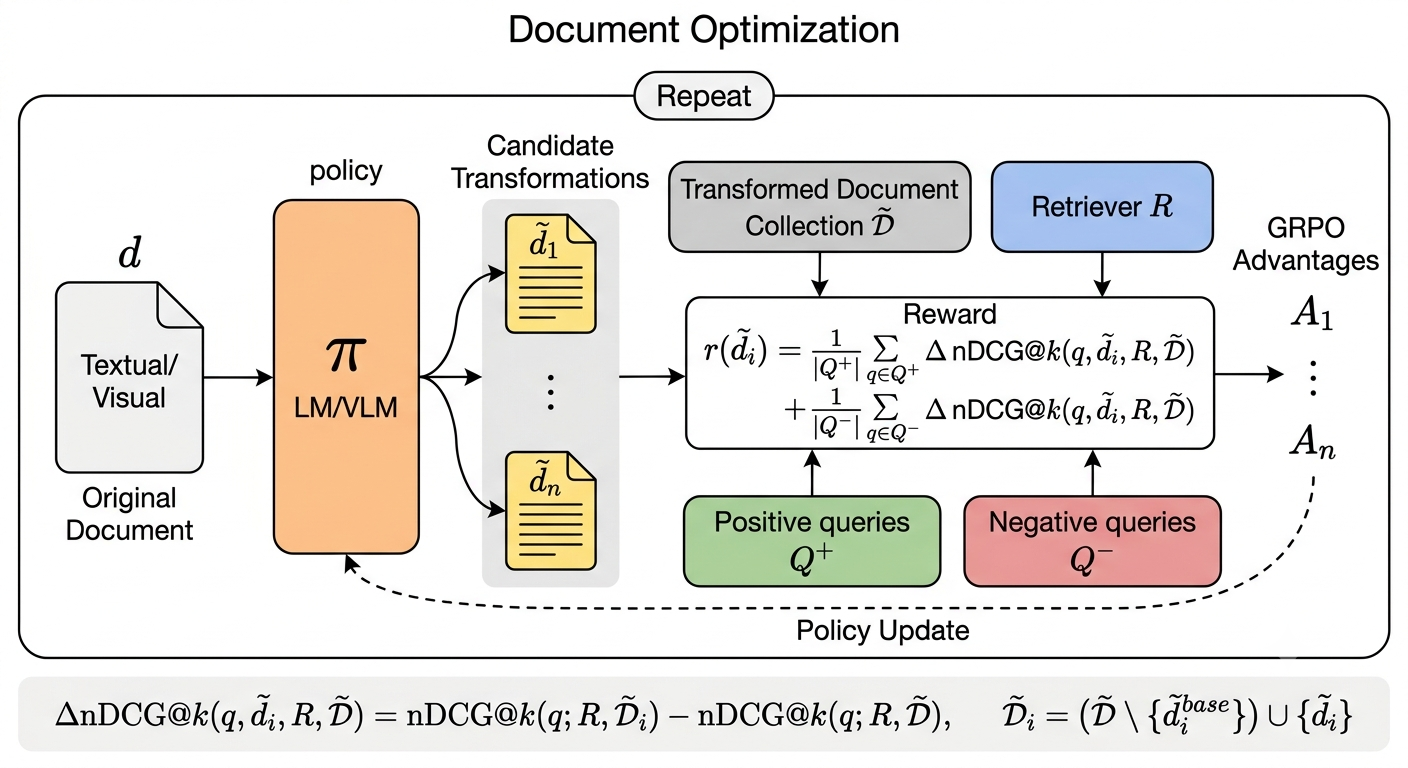}
    \caption{Document Optimization. At each iteration, a policy $\pi$ generates candidate transformations $\tilde d_i$ for a document $d$. Each transformation is evaluated under a fixed collection $\mathcal{\tilde D}$ and retriever $R$ using positive and negative queries, $Q^+$ and $Q^-$. Rewards are computed via the counterfactual $\Delta\mathrm{nDCG}@k$, capturing retrieval-quality gains on both positive and negative query sets. These are converted into GRPO advantages to update the policy. Through training, the policy learns to generate transformations which improve rankings under the retriever’s representation space. Bottom: definition of $\Delta\mathrm{nDCG}@k$.}
    \label{fig:method}
\end{figure}

Large language models (LLMs) have become central to modern information retrieval. While encoder models serve as the representational backbone of neural retrievers~\citep{lee-etal-2019-latent,karpukhin-etal-2020-dense,10.1145/3397271.3401075}, generative LLMs can further augment retrieval pipelines, for example by rewriting queries~\citep{10.1145/333135.333138,gao-etal-2023-precise,hsu2024groundingtryingllmsreinforcement} or reranking a limited set of candidate documents~\citep{sun-etal-2023-chatgpt,qin-etal-2024-large}. These applications, however, are constrained by latency and serving cost, as they place expensive LLM inference directly on the latency-critical retrieval path.

An alternative is to apply LLMs on the document side, where computation can be performed offline before indexing~\citep{nogueira2019documentexpansionqueryprediction,weller-etal-2024-generative,kuo2025doc2querytopiccoveragebaseddocument}. Rather than invoking an LLM for every query, each document can be transformed by an LLM for improved retrieval, allowing the cost to be amortized across all future queries while preserving a lightweight retrieval pipeline at inference time. However, naively transforming documents with generative LLMs does not reliably improve retrieval and can even degrade performance~\citep{10.1007/978-3-031-28238-6_31,weller-etal-2024-generative,kuo2025doc2querytopiccoveragebaseddocument}. LLMs are optimized with generative objectives, whereas retrieval is inherently discriminative: the value of a document rewrite lies in making the document more distinguishable from competing candidates for its target queries under the retriever’s notion of similarity.

In this paper, we introduce \emph{\textbf{Doc}ument \textbf{Opt}imization}, where we learn to rewrite documents into retrieval-oriented surrogates, under a specific retriever (\autoref{fig:method}), and show that it can be a powerful, flexible tool for improving retrieval in a variety of settings.

\begin{figure}[t]
    \centering
    \includegraphics[width=\linewidth,keepaspectratio]{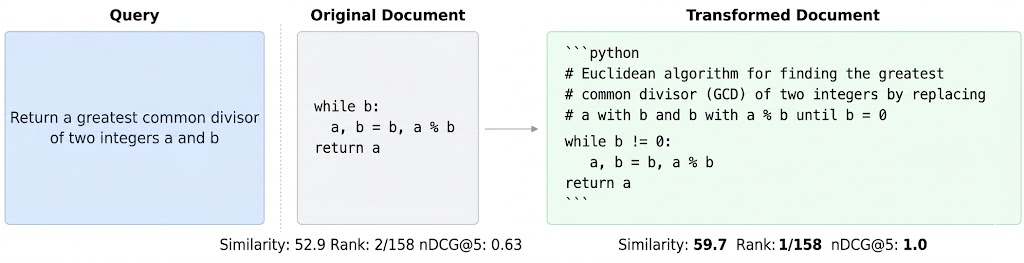}
    \caption{
    Example of document transformation improving retrieval quality. A HumanEval (MTEB) query (left) is matched against a labeled code document: the original (middle) and its transformed version (right). The transformation adds natural language comments, introduces minor code edits (e.g., \texttt{while b} $\rightarrow$ \texttt{while b != 0}), and wraps the code in a \texttt{python} block. Under the OpenAI text-embedding-3-small model, similarity increases (52.9 $\rightarrow$ 59.7), rank improves (2 $\rightarrow$ 1), and nDCG@5 rises (0.63 $\rightarrow$ 1.0).
    }
    \label{fig:code}
\end{figure}

Learning to rewrite documents is challenging because the objective is defined over discrete text, and the quality of a rewrite is only revealed through its downstream retrieval behavior. We therefore formulate document optimization as a reinforcement learning problem. In this mode, we treat an instruction-tuned language model (LM) or vision language model (VLM) as a policy over document rewrites. Rewards are defined counterfactually: for a proposed rewrite, we measure how retrieval would change if we replaced the original document in the index while all other documents were held fixed. Effective rewrites improve ranking of the document for relevant queries and suppress spurious matches to irrelevant ones. We optimize this policy using Group Relative Policy Optimization (GRPO) \citep{shao2024deepseekmathpushinglimitsmathematical}. For each document, the policy samples candidate rewrites and receives rewards based on their counterfactual retrieval utility, gradually learning to produce rewrites that are better aligned with the retriever’s representation space (\autoref{fig:code}).

This method applies in black-box settings and is compatible with single-vector~\citep{lee-etal-2019-latent,karpukhin-etal-2020-dense}, multi-vector~\citep{10.1145/3397271.3401075,santhanam-etal-2022-colbertv2}, and lexical retrievers~\citep{10.1561/1500000019} alike, since the learning signal depends only on retriever rankings, rather than gradients or access to retriever internals. We evaluate our approach on two retrieval settings: visual document retrieval, where images of PDF pages are converted into textual surrogates for text retrieval, and code retrieval, where code snippets are rewritten into retrieval-optimized textual representations. We test a diverse pool of retrievers, both open-weight and closed-weight, and compare our approach to using proprietary LLMs like GPT and Claude.

Across both visual document retrieval and code retrieval, we find that transforming documents with generative LLMs improves retrieval, and that optimizing these transformations with DocOpt yields additional gains. In the black-box setting, document optimization yields large consistent gains over direct retrieval and zero-shot transformation, for both closed-weight and open-weight models. For example, applying document optimization to OpenAI text-embedding-3-small improves nDCG@5 from 58.35→67.94 on average, both on code (68.63→78.23) and VDR (48.08→57.65). DocOpt outperforms both commercial LLM baselines, GPT-5.4-mini at 60.83 and Claude Haiku 4.5 at 63.75, and applying the 6.5$\times$ more expensive text-embedding-3-large at 66.15. We find that gains extend across families and architectures: Qwen3-Embedding-4B, improves from 63.97 to 71.17 on average, while Jina-ColBERT-V2 sees a substantial boost from 56.31 to 65.95. In the open-weight retriever setting, document optimization is competitive with fine-tuning on most settings, and combining both appears especially effective for multi-vector retrievers: Jina-ColBERT-V2 increases from 56.31 to 69.15, while finetuning alone only reaches 64.14.

Notably, all of these improvements are achieved entirely offline, preserving inference-time efficiency.

\section{Background}

\textbf{Passage Retrieval.} Let $D=(d_1,...,d_n)$ be a set of documents, and let $q$ denote a textual query. The retrieval task is to return the top-$k$ items of $D$ that are most relevant\footnote{Where relevance is defined by a binary label, typically via human annotation} to $q$, where $k$ is a hyperparameter. Retrieval quality is commonly evaluated using Recall@$k$ and nDCG@$k$. Recall@$k$ measures how many relevant documents appear in the top-$k$ results, and is particularly useful for evaluating first-stage retrieval in multi-stage pipelines. In this work we adopt nDCG@$k$ as the primary metric, since it additionally accounts for documents ranking in the top-$k$ by rewarding higher placements (see Appendix~\ref{app:ndcg_details} for formal definitions).

\textbf{Bi-Encoders.} Modern retrieval systems often rely on bi-encoders~\citep{reimers2019sentencebertsentenceembeddingsusing,karpukhin-etal-2020-dense,10.1145/3397271.3401075}, which map queries and documents independently into a shared representation space and score relevance through a non-parametric similarity function. These models are typically initialized from pretrained bidirectional LMs~\citep{devlin2019bertpretrainingdeepbidirectional} and adapted for retrieval using noise contrastive estimation~\citep{pmlr-v9-gutmann10a,oord2019representationlearningcontrastivepredictive} over labeled queries and documents. A key property of bi-encoders is that document representations can be precomputed offline. At inference time, the system encodes only the query and efficiently searches a prebuilt index~\citep{5432202,johnson2017billionscalesimilaritysearchgpus,10.1109/TPAMI.2018.2889473,santhanam2022plaidefficientenginelate}  to retrieve the top-$k$ highest-scoring documents.

\textbf{Foundation Models as Policies.}
An LM or a VLM can be described as a stochastic policy that maps from an input state (prompt) to a distribution over actions (next-token predictions), where a sequence of such predictions forms a trajectory. Policy gradient methods~\citep{10.1007/BF00992696,NIPS1999_464d828b,schulman2017proximalpolicyoptimizationalgorithms} have been 
effective at improving the performance of LMs on various tasks~\citep{ziegler2020finetuninglanguagemodelshuman,ouyang2022traininglanguagemodelsfollow,shao2024deepseekmathpushinglimitsmathematical}. More recently, Group Relative Policy Optimization (GRPO)~\citep{shao2024deepseekmathpushinglimitsmathematical} and its variants~\citep{liu2025understandingr1zeroliketrainingcritical,yu2025dapoopensourcellmreinforcement} have emerged as an effective approach, leveraging relative reward comparisons between action trajectories for variance reduction without requiring an explicit value function. In our approach, the policy model can be any generative model that maps inputs into the appropriate representation space for the retriever.
\section{Method}

We assume a supervised setting with access to annotated sets of positive and negative queries for each document $d_i$, denoted $Q_i^+$ and $Q_i^-$.

Given a retriever $R$, our goal is to learn a policy $\pi_\phi$ (in this section we will refer to the document optimization model as a policy) that generates a rewritten document $\tilde d$ such that $\tilde d$ improves retrieval performance under $R$ for queries relevant to $d$.

We initialize $\pi_\phi$ from an instruction-tuned LM or VLM, conditioned on a fixed prompt $\ptrans$ that specifies the desired transformation (we provide the prompts used in our experiments in Appendix~\ref{app:prompts}). We then sample a transformed document from the policy:
\begin{equation}
\tilde d \sim \pi_\phi(\cdot \mid \ptrans; d).
\end{equation}

\subsection{Counterfactual Retrieval Reward}

To evaluate a rewrite $\tilde d_i \sim \pi_\phi(\cdot \mid \ptrans; d_i)$, we measure its \emph{counterfactual impact} on retrieval under~$R$.

We first construct an initial rewritten collection by sampling a transformation for each document, yielding $\tilde{\mathcal{D}}$. For a given document $d_i$, we then define a counterfactual collection in which only its rewritten form is changed:
\begin{equation}
\tilde{\mathcal{D}}_i := (\tilde{\mathcal{D}} \setminus \{\tilde d_i^{\,\text{base}}\}) \cup \{\tilde d_i\},
\end{equation}
where $\tilde d_i^{\,\text{base}}$ denotes the baseline transformation of $d_i$ in $\tilde{\mathcal{D}}$. Importantly, $\tilde{\mathcal{D}}$ is held fixed when evaluating the effect of replacing $\tilde d_i^{\,\text{base}}$ with $\tilde d_i$.

Given a query $q$, let $\mathrm{nDCG@}k(q; R, \mathcal{D})$ denote the retrieval quality at cutoff $k$ (in our experiments we use $k=5$) when using retriever $R$ over collection $\mathcal{D}$. We define the counterfactual change for a query as:
\begin{equation}
\Delta \mathrm{nDCG@}k(q)
:=
\mathrm{nDCG@}k(q; R, \tilde{\mathcal{D}}_i)
-
\mathrm{nDCG@}k(q; R, \tilde{\mathcal{D}}).
\end{equation}

The reward aggregates these changes over positive and negative query sets $Q_i^{+}$ and $Q_i^{-}$ associated with $d_i$:
\begin{equation}
r_i(\tilde d_i) =
\underbrace{
\frac{1}{|Q_i^{+}|} \sum_{q \in Q_i^{+}}
\Delta \mathrm{nDCG@}k(q)
}_{\text{positive gain}}
+
\underbrace{
\frac{1}{|Q_i^{-}|} \sum_{q \in Q_i^{-}}
\Delta \mathrm{nDCG@}k(q)
}_{\text{negative gain}}.
\end{equation}

The baseline transformation $\tilde d_i^{\,\text{base}}$ can be interpreted as the current sample from the policy, and the reward measures how alternative samples compare to it under a fixed collection. This isolates the effect of the transformation on ranking, encouraging improved retrieval for relevant queries while discouraging spurious matches for negative queries. We only take training steps for documents with non-empty positive and negative query sets, avoiding division by zero in the reward computation.

\textbf{Reward Choice.} The reward directly optimizes the target evaluation metric, $\mathrm{nDCG@}k$, rather than relying on a proxy. It is agnostic to the underlying similarity function and applies uniformly across different retriever architectures. Notably, ranking-based rewards are sparse, as many document transformations yield identical scores when they do not affect the ranking. An alternative is to define the reward using query–document similarity, which provides a dense signal, but lacks the context of other documents in the index. While we find the ranking-based reward to perform better empirically, other formulations can also be effective under this framework. An empirical comparison is in \autoref{app:ablations}.

\subsection{Policy Optimization}

We optimize $\pi_\phi$ using Group Relative Policy Optimization (GRPO) \citep{shao2024deepseekmathpushinglimitsmathematical}. We provide the full details of the GRPO configuration we use in Appendix~\ref{app:grpo}. During each GRPO iteration, we sample a batch of documents $\mathcal{B} \subset D$ with annotated queries. For each document $d_i \in \mathcal{B}$, we generate $G$ candidate transformations from $\pi_\phi$, forming a group of candidate rewrites for that document. Each candidate is evaluated using the counterfactual retrieval reward against $\tilde D$, and we compute relative advantages within each group. We then perform a single policy update per batch, aggregating learning signals across all document-level groups. We use the temperature parameter to balance exploration and quality in the generated transformations. For index population, we use a low temperature to ensure stable, high-quality rewrites, while for generating the transformed collection during training, we use a higher temperature to encourage exploration. Finally, when training completes, we transform the entire document collection again using the learned policy and use the resulting corpus for retrieval. Computing rewards requires evaluating transformations against the retrieval index. However, this is computationally prohibitive for large $|D|$ at every step. Instead, we only re-transform the full collection periodically every $T_{\text{refresh}}$ steps, introducing a controlled degree of staleness. 

\subsection{Joint retriever-policy adaptation}
Document optimization is not a replacement for retriever fine-tuning, which can be highly beneficial when retriever weights are available. We therefore consider jointly adapting the retriever and the policy. In a straightforward variant, $R$ is periodically fine-tuned on the current transformed collection every $T_{\text{refresh}}$ steps. This can also be viewed as a form of data augmentation, where the retriever is trained on a different set of document transformations at each iteration.
\section{Experiments}

\begin{table}[t]
\centering
\small
\setlength{\tabcolsep}{4pt}
\caption{Retrieval results across code and visual domains. Deltas are absolute changes relative to Direct Retrieval. Mean weights the two domains equally. Visual Direct Retrieval uses OCR. Bold marks the best displayed value per retriever and column.}
\begin{adjustbox}{width=\columnwidth}
\begin{tabular}{llcccccc}
\toprule
\textbf{Retriever} & \textbf{Method} & \multicolumn{2}{c}{\textbf{Code}} & \multicolumn{2}{c}{\textbf{Visual}} & \multicolumn{2}{c}{\textbf{Mean}} \\
\cmidrule(lr){3-4}\cmidrule(lr){5-6}\cmidrule(lr){7-8}
 & & \textbf{val} & {\scriptsize$\Delta$} & \textbf{val} & {\scriptsize$\Delta$} & \textbf{val} & {\scriptsize$\Delta$} \\
\midrule
\multicolumn{8}{@{}l}{\textsc{Lexical}} \\
\midrule
\multirow{5}{*}{\hspace{0.5em}BM25} & \cellcolor{retrieverbg!85} Direct Retrieval & 15.23 & \deltacell{\zerodelta} & 14.13 & \deltacell{\zerodelta} & 14.68 & \deltacell{\zerodelta} \\
 & \cellcolor{basebg!85} GPT-5.4-mini & 37.90 & \deltacell{\pos{+22.67}} & 15.91 & \deltacell{\pos{+1.78}} & 26.90 & \deltacell{\pos{+12.22}} \\
 & \cellcolor{basebg!85} Claude Haiku 4.5 & 46.37 & \deltacell{\pos{+31.13}} & 18.25 & \deltacell{\pos{+4.13}} & 32.31 & \deltacell{\pos{+17.63}} \\
 & \cellcolor{basebg!85} Qwen-3 2B/4B & 53.03 & \deltacell{\pos{+37.80}} & 18.05 & \deltacell{\pos{+3.93}} & 35.54 & \deltacell{\pos{+20.86}} \\
 & \cellcolor{optbg!85} DocOpt & \multicolumn{1}{c}{\textbf{58.20}} & \deltacell{\pos{+42.97}} & \multicolumn{1}{c}{\textbf{19.93}} & \deltacell{\pos{+5.80}} & \multicolumn{1}{c}{\textbf{39.06}} & \deltacell{\pos{+24.38}} \\
\midrule
\multicolumn{8}{@{}l}{\textsc{Closed-weights}} \\
\midrule
\multirow{5}{*}{\hspace{0.5em}OpenAI text-emb-small} & \cellcolor{retrieverbg!85} Direct Retrieval & 68.63 & \deltacell{\zerodelta} & 48.08 & \deltacell{\zerodelta} & 58.35 & \deltacell{\zerodelta} \\
 & \cellcolor{basebg!85} GPT-5.4-mini & 70.84 & \deltacell{\pos{+2.20}} & 50.83 & \deltacell{\pos{+2.75}} & 60.83 & \deltacell{\pos{+2.48}} \\
 & \cellcolor{basebg!85} Claude Haiku 4.5 & 71.94 & \deltacell{\pos{+3.31}} & 55.57 & \deltacell{\pos{+7.49}} & 63.75 & \deltacell{\pos{+5.40}} \\
 & \cellcolor{basebg!85} Qwen-3 2B/4B & 75.13 & \deltacell{\pos{+6.50}} & 53.33 & \deltacell{\pos{+5.25}} & 64.23 & \deltacell{\pos{+5.88}} \\
 & \cellcolor{optbg!85} DocOpt & \multicolumn{1}{c}{\textbf{78.23}} & \deltacell{\pos{+9.60}} & \multicolumn{1}{c}{\textbf{57.65}} & \deltacell{\pos{+9.58}} & \multicolumn{1}{c}{\textbf{67.94}} & \deltacell{\pos{+9.59}} \\
\midrule
\multirow{5}{*}{\hspace{0.5em}OpenAI text-emb-large} & \cellcolor{retrieverbg!85} Direct Retrieval & 77.73 & \deltacell{\zerodelta} & 54.58 & \deltacell{\zerodelta} & 66.15 & \deltacell{\zerodelta} \\
 & \cellcolor{basebg!85} GPT-5.4-mini & 75.79 & \deltacell{\negd{-1.94}} & 55.81 & \deltacell{\pos{+1.23}} & 65.80 & \deltacell{\negd{-0.35}} \\
 & \cellcolor{basebg!85} Claude Haiku 4.5 & 77.22 & \deltacell{\negd{-0.52}} & 58.00 & \deltacell{\pos{+3.42}} & 67.61 & \deltacell{\pos{+1.45}} \\
 & \cellcolor{basebg!85} Qwen-3 2B/4B & 78.70 & \deltacell{\pos{+0.97}} & 57.03 & \deltacell{\pos{+2.45}} & 67.86 & \deltacell{\pos{+1.71}} \\
 & \cellcolor{optbg!85} DocOpt & \multicolumn{1}{c}{\textbf{81.87}} & \deltacell{\pos{+4.13}} & \multicolumn{1}{c}{\textbf{59.40}} & \deltacell{\pos{+4.83}} & \multicolumn{1}{c}{\textbf{70.63}} & \deltacell{\pos{+4.48}} \\
\midrule
\multicolumn{8}{@{}l}{\textsc{Open-weights}} \\
\midrule
\multirow{7}{*}{\hspace{0.5em}Qwen3-Emb-0.6B} & \cellcolor{retrieverbg!85} Direct Retrieval & 79.00 & \deltacell{\zerodelta} & 39.53 & \deltacell{\zerodelta} & 59.26 & \deltacell{\zerodelta} \\
 & \cellcolor{basebg!85} GPT-5.4-mini & 79.83 & \deltacell{\pos{+0.83}} & 46.92 & \deltacell{\pos{+7.39}} & 63.37 & \deltacell{\pos{+4.11}} \\
 & \cellcolor{basebg!85} Claude Haiku 4.5 & 80.07 & \deltacell{\pos{+1.07}} & 37.34 & \deltacell{\negd{-2.19}} & 58.70 & \deltacell{\negd{-0.56}} \\
 & \cellcolor{basebg!85} Qwen-3 2B/4B & 80.70 & \deltacell{\pos{+1.70}} & 52.53 & \deltacell{\pos{+13.00}} & 66.61 & \deltacell{\pos{+7.35}} \\
 & \cellcolor{optbg!85} DocOpt & 81.93 & \deltacell{\pos{+2.93}} & \multicolumn{1}{c}{\textbf{57.35}} & \deltacell{\pos{+17.83}} & \multicolumn{1}{c}{\textbf{69.64}} & \deltacell{\pos{+10.38}} \\
 & \cellcolor{ftbg!85} Retriever Fine-tuning & 80.87 & \deltacell{\pos{+1.87}} & 56.28 & \deltacell{\pos{+16.75}} & 68.57 & \deltacell{\pos{+9.31}} \\
 & \cellcolor{jointbg!85} DocOpt - Joint Adaptation & \multicolumn{1}{c}{\textbf{82.50}} & \deltacell{\pos{+3.50}} & 54.33 & \deltacell{\pos{+14.80}} & 68.41 & \deltacell{\pos{+9.15}} \\
\midrule
\multirow{7}{*}{\hspace{0.5em}Qwen3-Emb-4B} & \cellcolor{retrieverbg!85} Direct Retrieval & 83.43 & \deltacell{\zerodelta} & 44.50 & \deltacell{\zerodelta} & 63.97 & \deltacell{\zerodelta} \\
 & \cellcolor{basebg!85} GPT-5.4-mini & 81.31 & \deltacell{\negd{-2.12}} & 47.83 & \deltacell{\pos{+3.33}} & 64.57 & \deltacell{\pos{+0.61}} \\
 & \cellcolor{basebg!85} Claude Haiku 4.5 & 81.67 & \deltacell{\negd{-1.77}} & 43.39 & \deltacell{\negd{-1.11}} & 62.53 & \deltacell{\negd{-1.44}} \\
 & \cellcolor{basebg!85} Qwen-3 2B/4B & 82.97 & \deltacell{\negd{-0.47}} & 56.25 & \deltacell{\pos{+11.75}} & 69.61 & \deltacell{\pos{+5.64}} \\
 & \cellcolor{optbg!85} DocOpt & \multicolumn{1}{c}{\textbf{84.63}} & \deltacell{\pos{+1.20}} & 57.70 & \deltacell{\pos{+13.20}} & 71.17 & \deltacell{\pos{+7.20}} \\
 & \cellcolor{ftbg!85} Retriever Fine-tuning & 84.17 & \deltacell{\pos{+0.73}} & \multicolumn{1}{c}{\textbf{59.23}} & \deltacell{\pos{+14.73}} & \multicolumn{1}{c}{\textbf{71.70}} & \deltacell{\pos{+7.73}} \\
 & \cellcolor{jointbg!85} DocOpt - Joint Adaptation & 84.20 & \deltacell{\pos{+0.77}} & 57.98 & \deltacell{\pos{+13.48}} & 71.09 & \deltacell{\pos{+7.12}} \\
\midrule
\multirow{7}{*}{\hspace{0.5em}Jina-Colbert-V2} & \cellcolor{retrieverbg!85} Direct Retrieval & 58.87 & \deltacell{\zerodelta} & 53.75 & \deltacell{\zerodelta} & 56.31 & \deltacell{\zerodelta} \\
 & \cellcolor{basebg!85} GPT-5.4-mini & 68.67 & \deltacell{\pos{+9.80}} & 52.81 & \deltacell{\negd{-0.94}} & 60.74 & \deltacell{\pos{+4.43}} \\
 & \cellcolor{basebg!85} Claude Haiku 4.5 & 71.02 & \deltacell{\pos{+12.15}} & 53.05 & \deltacell{\negd{-0.71}} & 62.03 & \deltacell{\pos{+5.72}} \\
 & \cellcolor{basebg!85} Qwen-3 2B/4B & 70.50 & \deltacell{\pos{+11.63}} & 55.83 & \deltacell{\pos{+2.07}} & 63.16 & \deltacell{\pos{+6.85}} \\
 & \cellcolor{optbg!85} DocOpt & 73.87 & \deltacell{\pos{+15.00}} & 58.03 & \deltacell{\pos{+4.27}} & 65.95 & \deltacell{\pos{+9.64}} \\
 & \cellcolor{ftbg!85} Retriever Fine-tuning & 71.53 & \deltacell{\pos{+12.67}} & 56.75 & \deltacell{\pos{+3.00}} & 64.14 & \deltacell{\pos{+7.83}} \\
 & \cellcolor{jointbg!85} DocOpt - Joint Adaptation & \multicolumn{1}{c}{\textbf{75.00}} & \deltacell{\pos{+16.13}} & \multicolumn{1}{c}{\textbf{63.30}} & \deltacell{\pos{+9.55}} & \multicolumn{1}{c}{\textbf{69.15}} & \deltacell{\pos{+12.84}} \\
\bottomrule
\end{tabular}
\end{adjustbox}
\label{tab:combined_doc_retrieval}
\end{table}

\subsection{Setup}
\textbf{Tasks.} We evaluate our method on two retrieval settings: visual document retrieval (VDR) and code retrieval. VDR \citep{mathew2021docvqadatasetvqadocument,mathew2021infographicvqa,li-etal-2024-multimodal-arxiv,zhu2022towards,faysse2025colpali} considers a collection of visually rich documents, including charts, images, and tables, paired with text queries grounded in the document content. We apply our method to perform image-to-text transformations of visual content. In this setting, the transformation policy maps each document image into a textual surrogate that is subsequently embedded by a text retriever. Code Retrieval assumes a corpus of code documents (e.g., functions or code snippets) paired with natural language queries describing their functionality. Here, the policy rewrites each code snippet into a retrieval-oriented textual representation (e.g., \autoref{fig:code}), making the document easier to match with natural-language 
queries.

These domains are natural candidates for testing the value of LLMs for document-side adaptation. Converting PDFs to text is a common practice which already constitutes a form of document transformation. In code retrieval, code snippets can be enriched with comments or rewritten into more readable formats that could make them easier to retrieve. For evaluation, we use four datasets from ViDoRe2~\citep{mace2025vidore} for VDR, and code retrieval tasks from MTEB~\citep{muennighoff2023mtebmassivetextembedding}: HumanEvalRetrieval~\citep{chen2021codex},MBPPRetrieval~\citep{austin2021programsynthesislargelanguage}, and DS1000Retrieval~\citep{lai2022ds1000naturalreliablebenchmark}.

\textbf{Retrievers.} For black-box retrievers, we use OpenAI's text-embedding-3-small and text-embedding-3-large through the OpenAI API. For open-weights models, we use Qwen3-Embedding-0.6B and Qwen3-Embedding-4B as single-vector retrievers, and Jina-ColBERT-V2 as a multivector one. We additionally evaluate BM25 as a lexical retriever. 

\textbf{Generative models.} For visual document retrieval, we use Qwen3-VL-2B-Instruct~\citep{qwen3technicalreport} to generate textual descriptions from document images. For code, we use Qwen3-4B-Instruct-2507~\citep{qwen3technicalreport}. We additionally evaluate two proprietary LLMs, GPT-5.4 mini~\citep{openai2026gpt54mini} and Claude Haiku 4.5~\citep{anthropic2025haiku45}, as zero-shot document rewriters. 

\textbf{Baselines.} For each task, we compare DocOpt against (1) \textit{direct retrieval} without document transformation. In the visual setting, we use an OCR model as the direct retrieval baseline, as we focus only on text retrieval. (2) \textit{zero-shot document transformations} without any training, via proprietary LLMs or the Qwen checkpoint we initiate DocOpt from. In open-weight retriever settings, we additionally evaluate (3) \textit{retriever fine-tuning} and (4) \textit{DocOpt - joint adaptation} of the retriever and transformation policy.

\textbf{Evaluation.}
We use nDCG@5 as our evaluation metric. We use a small train split for finetuning and training (exact details are provided in Appendix \ref{app:datasets}). The \textit{direct retrieval} and \textit{zero-shot transformation} baselines are applied to the test queries directly, and the \textit{policy optimization}, \textit{retriever fine-tuning} and \textit{joint adaptation} methods are first trained on the train split and then evaluated similarly on the test set. To ensure a fair comparison, we allocate equal tuning effort to all three.

\subsection{Results}
\autoref{tab:combined_doc_retrieval} shows that document transformation improves retrieval quality across retrievers and datasets. Zero-shot rewriting already leads to retrieval gains over direct retrieval, with larger gains for Qwen-3 models over commerical API models. DocOpt in turn yields further gains, outperforming direct retrieval and zero-shot transformation on average across all retrievers (\autoref{app:per-domain-res} details results per dataset in each domain). It demonstrates the benefit of not merely using an LLM to produce richer document representations, but to explicitly optimize those representations for the target retriever. For example, with BM25, DocOpt reaches 39.06 mean nDCG@5, compared with 26.90 for GPT-5.4-mini and 32.31 for Claude Haiku 4.5; with OpenAI text-embedding-3-small, it reaches 67.94, compared with 60.83 and 63.75, respectively. 

In the black-box setting, DocOpt yields substantial gains for both OpenAI retrievers and across both domains. For text-embedding-3-small, it improves mean nDCG@5 from 58.35 to 67.94, corresponding to nearly identical gains on code (68.63$\rightarrow$78.23) and VDR (48.08$\rightarrow$57.65). Notably, the optimized small model even surpasses direct retrieval with the 6.5$\times$ more expensive text-embedding-3-large model (66.15 mean nDCG@5). DocOpt also improves the larger retriever itself, from 66.15 to 70.63 on average, showing that document optimization provides gains even when the underlying retriever is already strong.

In the open-weight setting, DocOpt remains competitive with directly fine-tuning the retriever, despite modifying only the document representations. It outperforms retriever fine-tuning for Qwen3-Embedding-0.6B (69.64 vs.\ 68.57) and Jina-ColBERT-V2 (65.95 vs.\ 64.14), while closely matching it for Qwen3-Embedding-4B (71.17 vs.\ 71.70). The gains are particularly large on VDR for the Qwen retrievers, where DocOpt improves Qwen3-Embedding-0.6B by 17.83 points and Qwen3-Embedding-4B by 13.20 points, and on code for Jina-ColBERT-V2, where it improves performance by 15.00 points. Moreover, document-side and retriever-side adaptation can be complementary: jointly optimizing both yields the strongest result for Jina-ColBERT-V2, reaching 69.15 mean nDCG@5 compared with 65.95 for DocOpt alone and 64.14 for retriever fine-tuning alone.
\section{Analysis}

\paragraph{End-to-End Costs.}
\label{sec:offline-cost}

We use vLLM~\citep{kwon2023efficientmemorymanagementlarge} for corpus rewriting and Hugging Face TRL for GRPO training. Our cost comparison
uses Qwen3-4B-Instruct-2507 on DS1000Retrieval~\citep{lai2022ds1000naturalreliablebenchmark}, whose documents average 183 source tokens. We time rewriting over all 1,998 documents per method and extrapolate to 10K and 1M documents.

On our hardware, policy optimization runs averaged 13.87/10.17 seconds per step on a single A100/H100. Training uses 407 queries against a fixed 1,998-document index; the 150-step budget includes full-index refreshes after steps 50 and 100. We reuse the same policy at both deployment scales.
This gives 0.54 H100-hours, including approximately 6.9 minutes of initialization, refresh, and other overhead. Preparation costs \$1.95, including \$0.07 for retriever embeddings.

API costs use official Batch pricing. DocOpt GPU time is valued at \$3.49/H100-hour. Target-index
construction and unrecorded query-generation time are excluded.
As \autoref{tab:costs} shows, DocOpt costs \$2.12 at 10K documents,
below all API baselines. At 1M documents, its preparation and
rewriting cost approximately $33\times$ less and run $2.9\times$
faster than GPT-5.4 mini (\$18.59 vs.\ \$607.49;
5.31 vs.\ 15.24 hours).

\begin{table}[t]
\centering
\caption{Estimated policy-preparation and rewriting time and cost
at 10K and 1M documents, using a 150-step policy budget.
DocOpt is cheapest at both scales and fastest at 1M.
Totals are calculated before rounding.}
\label{tab:costs}
\small
\setlength{\tabcolsep}{4pt}
\begin{tabular}{llrrrrrr}
\toprule
Docs & Method
& \multicolumn{2}{c}{Policy preparation}
& \multicolumn{2}{c}{Rewriting}
& \multicolumn{2}{c}{Total} \\
\cmidrule(lr){3-4}
\cmidrule(lr){5-6}
\cmidrule(lr){7-8}
& & Time (h) & Cost (\$)
  & Time (h) & Cost (\$)
  & Time (h) & Cost (\$) \\
\midrule
10K & DocOpt (H100)
    & 0.54 & 1.95 & 0.05 & 0.17 & 0.59 & \textbf{2.12} \\
    & GPT-5.4 mini
    & --- & --- & 0.15 & 6.07 & \textbf{0.15} & 6.07 \\
    & Claude Haiku 4.5
    & --- & --- & 0.26 & 9.74 & 0.26 & 9.74 \\
    & GPT-5.4
    & --- & --- & 0.33 & 21.01 & 0.33 & 21.01 \\
    & Claude Sonnet 4.6
    & --- & --- & 0.80 & 34.55 & 0.80 & 34.55 \\
\midrule
1M  & DocOpt (H100)
    & 0.54 & 1.95 & 4.77 & 16.64
    & \textbf{5.31} & \textbf{18.59} \\
    & GPT-5.4 mini
    & --- & --- & 15.24 & 607.49 & 15.24 & 607.49 \\
    & Claude Haiku 4.5
    & --- & --- & 26.37 & 974.36 & 26.37 & 974.36 \\
    & GPT-5.4
    & --- & --- & 33.06 & 2,101.32 & 33.06 & 2,101.32 \\
    & Claude Sonnet 4.6
    & --- & --- & 79.90 & 3,454.62 & 79.90 & 3,454.62 \\
\bottomrule
\end{tabular}
\end{table}

\begin{figure}[tb]
    \centering
    \includegraphics[width=0.6\linewidth]{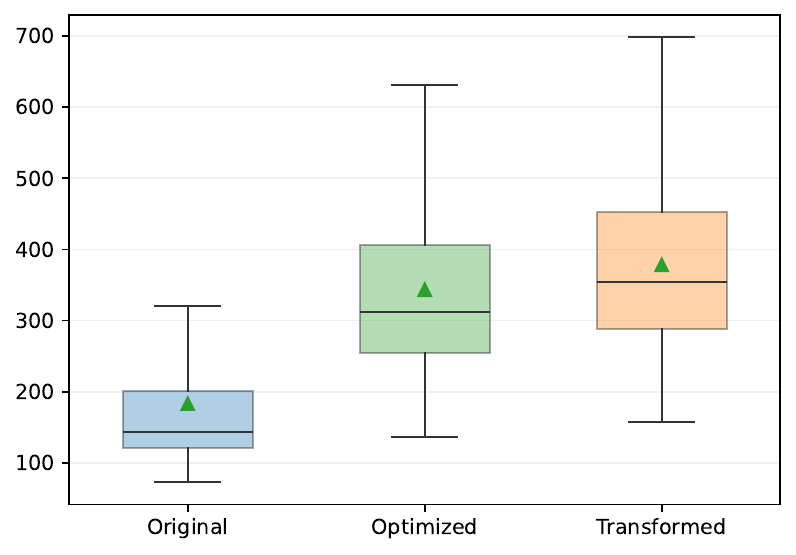}
    \caption{
    Distribution of document lengths (in tokens) for original, optimized documents and (zero-shot) transformed, on the DS1000 dataset using Qwen3-Embedding-4B as the retriever.
    }
    \label{fig:length_tokens}
\end{figure}

\paragraph{Qualitative Analysis.} We conduct qualitative analysis using Qwen3-Embedding-4B as the retriever on the DS10K dataset. The baseline retrieval performance on the original dataset is 61.2 nDCG@5, which remains largely unchanged after zero-shot transformation (61.1), but improves to 63.8 following document optimization. We analyze the distribution of document lengths across original, zero-shot transformed, and optimized documents and provide the statistics here (also visualized in \autoref{fig:length_tokens}). Original documents have a mean of 183 tokens and a median of 143. Zero-shot transformations substantially increase length, with a mean of 382 and a median of 355, nearly doubling document size. In contrast, optimized documents are more compact than zero-shot rewrites, with a mean of 346 and a median of 311. This indicates that the learned policy does not simply expand content, but selectively refines it to retain useful information, consistent with the goal of our approach.

\begin{figure}[t]
    \centering
    \includegraphics[width=\linewidth]{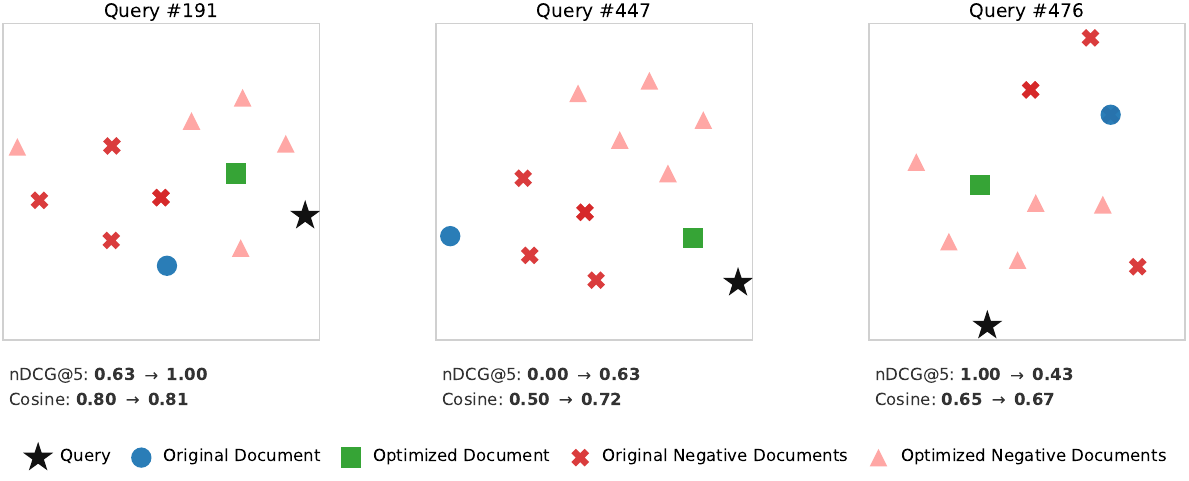}
    \caption{Two-dimensional t-SNE projections of Qwen3-Embedding-4B embeddings across different queries.}
    \label{fig:small_embs}
\end{figure}
 
\autoref{fig:small_embs} visualizes embeddings of queries, original documents, optimized documents, and negatives using a two-dimensional t-SNE projection
for qualitative inspection (additional examples are provided in \autoref{fig:embs} in the appendix). While 2D distances do not faithfully reflect similarities in the original space, they provide a
useful approximation of the local geometry. We observe that optimized documents form
clusters and in many cases where optimization improves nDCG@5, shift closer to the query and farther from negatives. Importantly, improvements are not solely driven by reduced query–document distance, but by favorable relative positioning with respect to competing negatives, consistent with the ranking-based reward.

\begin{table*}[t]
\caption{nDCG@5 on BRIGHT using Qwen3-Embedding-0.6B and OpenAI text-embedding-3-small. DocOpt achieves the best average retrieval performance with both retrievers, with larger gains for Qwen3-Embedding-0.6B.}
\centering
\small
\resizebox{\textwidth}{!}{%
\begin{tabular}{lc|*{7}{c}}
\toprule
Method & Avg. & Bio. & Earth & Econ. & Psych. & Robot. & Sust. & SO \\
\midrule
\multicolumn{9}{l}{\textit{Qwen3-Embedding-0.6B}} \\
\midrule
Direct Retrieval
& 13.06 & 10.63 & 19.92 & 15.28 & 14.31 & 11.28 & 9.98 & \textbf{9.99} \\
Zero-shot Transformation
& 12.72 & 11.70 & 23.78 & 14.58 & 14.80 & 9.41 & 8.72 & 6.08 \\
DocOpt
& \textbf{14.63} & \textbf{11.88} & \textbf{25.05} & \textbf{18.20} & \textbf{15.32} & \textbf{11.73}
& \textbf{11.14} & 9.08 \\
\midrule
\multicolumn{9}{l}{\textit{OpenAI text-embedding-3-small}} \\
\midrule
Direct Retrieval
& 16.47 & \textbf{20.10} & 27.01 & \textbf{17.18} & \textbf{22.00} & \textbf{9.95} & 12.98 & 6.06 \\
Zero-shot Transformation
& 15.45 & 16.88 & 29.94 & 15.78 & 19.12 & 8.42 & 12.52 & 5.49 \\
DocOpt
& \textbf{16.70} & 18.32 & \textbf{31.74} & 16.70 & 17.87 & 9.38 & \textbf{14.12} & \textbf{8.76} \\
\bottomrule
\end{tabular}%
}
\label{tab:bright-retrieval}
\end{table*}

\paragraph{Large Scale Natural Language Retrieval.} We additionally test DocOpt on a large-scale natural language retrieval setting. We use the StackExchange subset of the BRIGHT benchmark~\citep{su2025bright}for evaluation, which contains 7 datasets spanning diverse domains with over 500K documents in total. \autoref{tab:bright-retrieval} shows that DocOpt remains effective across both retrievers. With Qwen3-Embedding-0.6B, DocOpt improves average nDCG@5 by 12.0\% over direct retrieval and 15.0\% over zero-shot transformation, outperforming direct retrieval on 6 of 7 domains and zero-shot transformation on all 7. With OpenAI text-embedding-3-small, gains are more modest: DocOpt improves the average by 1.4\% over direct retrieval and 8.1\% over zero-shot transformation, outperforming them on 3 and 6 of 7 domains, respectively.

\paragraph{Transfer across Retrievers.} We test whether document transformations learned for one retriever transfer to another retriever. We run DocOpt against Qwen3-Embedding-0.6B and against OpenAI text-embedding-3-small, and test how well the rewriting policies generalize to other retrievers. \autoref{tab:vidore-retriever-transfer} illustrates that DocOpt transfers more effectively within retriever families. Qwen-trained rewrites improve Qwen 4B by 1.40 nDCG@5 points, while OpenAI-trained rewrites improve OpenAI large by 1.53 points. Across-family transfer is less consistent: Qwen-trained rewrites improve OpenAI large by 0.71 points, whereas it reduces OpenAI-small by 1.14 and OpenAI-trained rewrites reduce Qwen 4B performance by 0.30 points.

\begin{table}[t]
\caption{Transfer across retrievers on ViDoRe2. DocOpt transfers most effectively within retriever families, while cross-family transfer is less consistent.}
\centering\small
\begin{tabular}{lrrrr}
\toprule
Rewrite Policy & Qwen 0.6B & Qwen 4B & OpenAI small & OpenAI large \\
\midrule
No training (Baseline) & 52.53 & 56.25 & 53.33 & 57.03 \\
Qwen-trained & \textbf{57.35} & \textbf{57.65} & 52.19 & 57.74 \\
OpenAI-trained & 51.52 & 55.95 & \textbf{57.65} & \textbf{58.56} \\
\bottomrule
\end{tabular}
\label{tab:vidore-retriever-transfer}
\end{table}

\paragraph{Training on Synthetic queries.}

\begin{table}[t]
\caption{Learning from synthetic queries on ViDoRe2. Synthetic-query supervision improves average retrieval over zero-shot rewriting for both retrievers, but remains below labeled-query supervision.}
\centering\small
\begin{tabular}{lrrrrr}
\toprule
Supervision & Bio. & Econ. & ESG & ESG-H & Mean \\
\midrule
\multicolumn{6}{l}{\textit{Qwen3-Embedding-0.6B}} \\
Zero-shot & 54.40 & 45.80 & 55.80 & 54.10 & 52.53 \\
Labeled queries & 58.50 & 53.90 & 58.80 & 58.20 & 57.35 \\
Synthetic queries & 55.70 & 46.10 & 55.30 & 54.10 & 52.80 \\
\midrule
\multicolumn{6}{l}{\textit{OpenAI text-embedding-3-small}} \\
Zero-shot & 53.80 & 47.00 & 56.90 & 55.60 & 53.33 \\
Labeled queries & 55.50 & 51.40 & 61.80 & 61.90 & 57.65 \\
Synthetic queries & 54.60 & 48.40 & 56.30 & 57.70 & 54.25 \\
\bottomrule
\end{tabular}
\label{tab:vidore-synthetic}
\end{table}

\autoref{tab:vidore-synthetic} shows synthetic-query supervision improves average retrieval over zero-shot rewriting for both retrievers, while remaining below labeled-query supervision. Gains vary across collections and are larger for OpenAI small, with slight declines on ESG for both retrievers. \autoref{app:synthetic} provides the implementation details.

\section{Ablations and Design Choices}
\label{app:ablations}
\begin{table}[t]
\centering
\small
\begin{adjustbox}{width=\columnwidth}
\begin{tabular}{llcccc}
\toprule
\textbf{Type} & \textbf{Setting} 
& \textbf{HumanEval} 
& \textbf{MBPP} 
& \textbf{DS10K} 
& \textbf{AVG} \\

\midrule
Zero-shot Transformation & &  93.6 & 84.0 & 47.8 & 75.1 \\

\midrule

\multirow{2}{*}{\textbf{Similarity}} 
& Positives Only & 94.5 & \textbf{84.8} & 49.1 & 76.1 \\
& Positives + Negatives & 93.1  & \textbf{84.8 }& 51.3 & 76.4 \\

\midrule

\multirow{2}{*}{\textbf{Ranking}} 
& Positives Only &  92.9 & 84.2 & 50.9 & 76.0 \\
& Positives + Negatives & \textbf{94.9} & \textbf{84.8} & \textbf{52.1} & \textbf{77.3} \\

\bottomrule
\end{tabular}
\end{adjustbox}
\caption{Comparison of reward function variants.}
\label{tab:reward_function_choice}
\end{table}
\begin{table}[t]
\centering
\begin{tabular}{lccccc}
\toprule
\textbf{Setting} 
& \textbf{\#Queries}
& \textbf{HumanEval} 
& \textbf{MBPP} 
& \textbf{DS10K} 
& \textbf{AVG} \\

\midrule
Zero-shot Transformation & &  93.6 & 84.0 & 47.8 & 75.1 \\

\midrule
Random Sampling & 1  & 94.3 & 84.4 & 51.5 & 76.7 \\
Hard Negatives & 1   & 94.7 & 84.1 & \textbf{52.2} & 77.0 \\

\midrule
Random Sampling & 5  & 92.9 & 83.9 & 51.4 & 76.1 \\
Hard Negatives & 5   & 94.9 & \textbf{84.8} & 52.1 & \textbf{77.3} \\

\midrule
Random Sampling & 10 & \textbf{95.1} & 83.3 & 51.3 & 76.6 \\
Hard Negatives & 10  & 94.9 & \textbf{84.8} & 51.7 & 77.1 \\

\bottomrule
\end{tabular}
\caption{Comparison of weak supervision strategies for negative queries.}
\label{tab:weak_supervision}
\end{table}

We analyze key design choices in our approach, focusing on the reward formulation and negative query construction. All ablations are conducted on three code retrieval datasets with OpenAI text-embedding-3-small as the retriever, using a reduced optimization budget but otherwise the same training protocol as in the main experiments. Design choices were made on a separate validation split drawn from the training queries; for comparability with the main results, the tables report the same configurations evaluated on the test queries, which were not used for any design decisions.

\textbf{Reward function choice.}
\label{subsec:reward}
\autoref{tab:reward_function_choice}  shows that all learned reward variants improve over the zero-shot baseline, supporting document optimization with reinforcement learning as a general framework. The strongest overall results are achieved by the ranking-based reward that uses both positive and negative queries. Negative queries consistently help under both ranking-based and similarity-based objectives, indicating that discouraging spurious irrelevant matches is important for learning effective document transformations.

\textbf{Negative Queries.}
\label{subsec:weak_supervision}
\autoref{tab:weak_supervision} compares strategies for constructing negative queries, including random sampling and hard negative mining, as well as varying the number of negatives per document. We find that hard negatives outperform random sampling. Increasing the number of negatives yields diminishing returns, with the strongest results obtained using 5 hard negatives.

\section{Related Work}

\textbf{Document-side adaptation.} Document expansion~\citep{10.1145/312624.312681,tao-etal-2006-language,10.1145/1871437.1871571,10.1145/2348283.2348405} is a longstanding technique in information retrieval, designed to address the vocabulary mismatch problem~\citep{10.1145/32206.32212} in lexical retrieval. Later expansion solutions leverage neural models to predict likely queries and augment documents accordingly~\citep{nogueira2019documentexpansionqueryprediction,Nogueira2019FromDT,10.1007/978-3-031-28238-6_31,kuo2025doc2querytopiccoveragebaseddocument}. None of these treat document adaptation as an optimization problem, nor do they directly optimize it for retrieval performance. Closest to our work, \citet{liu2025harnessing} jointly optimize query and document rewriting with reinforcement learning, and find only modest gains from document-only optimization. However, their document-only conclusion is based on NFCorpus, where queries are extremely short (median length of two words), which may underrepresent settings with more nuanced information needs. In contrast, we focus on document-side optimization that incurs no additional query-time cost, across more diverse retrieval settings.

\textbf{Query-side methods and multi-stage retrieval. }Many lines of work improve retrieval by modifying the query or introducing additional stages at inference time. Query expansion~\citep{10.5555/188490.188508,10.1145/312624.312681,10.1145/333135.333138} and pseudo-relevance feedback~\citep{10.1145/3459637.3482124,sung-etal-2023-optimizing} augment the input query with additional terms, while more recent approaches leverage language models for query rewriting~\citep{gao-etal-2023-precise,hsu2024groundingtryingllmsreinforcement}. In parallel, multi-stage retrieval pipelines~\citep{kuzi2020leveragingsemanticlexicalmatching,10.1007/978-3-030-99736-6_7,uzan2025guidedqueryrefinementmultimodal} combine an initial retrieval step with reranking models to refine results. While these methods can substantially improve retrieval quality, they incur additional query-time cost and system complexity.

\textbf{Visual Document Retrieval. }Existing approaches can be broadly divided into image-based and text-based methods. Single-vector image-based models embed page images and queries into a shared embedding space~\citep{radford2021learningtransferablevisualmodels}, while more recent architectures adopt multi-vector representations to capture fine-grained interactions~\citep{faysse2025colpali}. In contrast, text-based approaches convert documents into textual surrogates using OCR or vision-language models~\citep{berrios2023languagemodelsseecomputer,nguyen-etal-2025-serval}, and leverage the mature infrastructure and ecosystem of text retrieval. We focus on this setting; however, unlike prior work that treats the surrogate heuristically, we optimize it to better align with the target retriever.

\section{Conclusion}
We presented document optimization as a framework for improving retrieval by rewriting documents to better align with a target retriever. Our method requires only black-box access, applies across retrieval architectures, and improves performance on both visual and code retrieval tasks. These results suggest that optimizing the document space itself is a promising and underexplored direction. We encourage future work to extend the framework to other domains and tasks, and further study the interplay between document-side optimization and model-side adaptation.




\bibliography{iclr2027_conference}
\bibliographystyle{iclr2027_conference}

\appendix
\appendix
\section{Appendix}

\subsection{Retrieval Metrics}
\label{app:ndcg_details}

Let $y_i \in \{0,1\}$ denote the relevance of the document at rank $i$.

\paragraph{Recall@$k$.}
\begin{equation}
\mathrm{Recall@}k = \frac{\sum_{i=1}^{k} y_i}{\sum_{i=1}^{n} y_i}.
\end{equation}

\paragraph{nDCG@$k$.}
\begin{equation}
\mathrm{nDCG@}k = \frac{\sum_{i=1}^{k} \frac{y_i}{\log_2(i + 1)}}{\max_{\pi} \sum_{i=1}^{k} \frac{y_{\pi(i)}}{\log_2(i + 1)}},
\end{equation}
where the denominator corresponds to the ideal (best possible) ranking.

\subsection{Document Transformation Prompts}
\label{app:prompts}
\paragraph{Visual Document Transformation Prompt}
\begin{verbatim}
Provide a comprehensive description of the document in the image in English.

Begin with a summary, then follow with details.

Extract all visible text and numerical values from the document.
\end{verbatim}
Also used in \cite{nguyen-etal-2025-serval}

\paragraph{Code Transformation Prompt}
\begin{verbatim}
Improve this code snippet.

You may:
- Add a short natural-language summary.
- Add comments where helpful.
- Rename variables or refactor for readability.

STRICT RULES:
- DO NOT change the behavior or functionality of the source document.
- Preserve the exact input-output semantics.
- Do not add facts, assumptions, or code behavior that is not
supported by the source.
- Keep the output concise and high-signal.

Return only the updated code with comments/documentation.
Do not add text outside the code.
\end{verbatim}

\subsection{GRPO Configuration}
\label{app:grpo}
We use the TRL library for optimization. We train in BF16 precision, without LoRA. We run each training run for 150 steps, and cap the rollouts at 1024 tokens.
We use these hyperparameters:

\begin{itemize}
\item \texttt{index\_refresh\_rate} = 50
\item \texttt{batch\_size} = 32
\item \texttt{num\_generations} = 8
\item \texttt{learning\_rate} = \(1\times10^{-5}\)
\item \texttt{beta} = 0.1
\item \texttt{max\_document\_tokens} = 1024,
\item \texttt{max\_prompt\_length} = 4096,
\item \texttt{temperature} = 0.7,
\item \texttt{top\_p} = 0.8,
\item \texttt{top\_k} = 20,
\item \texttt{repetition\_penalty} = 1.0,
\item \texttt{presence\_penalty} = 0
\end{itemize}

\subsection{Models}
We use a diverse set of retrievers and language models to evaluate our approach across architectures and access settings. Table~\ref{tab:appendix_retrievers} lists lexical and neural retrievers, including both open- and closed-weight models, as well as single- and multi-vector variants. Table~\ref{tab:appendix_lms} summarizes the instruction-tuned language models used to parameterize the transformation policy, covering both text-only and vision-language settings.

\begin{table}[H]
\centering
\small
\setlength{\tabcolsep}{6pt}
\renewcommand{\arraystretch}{1.15}

\begin{adjustbox}{max width=\linewidth}
\begin{tabular}{p{4.6cm} p{3.8cm} c c c}
\toprule
\textbf{Retriever} & \textbf{Weights} & \textbf{Params} & \textbf{Dim} & \textbf{Vector Type} \\
\midrule
\texttt{BM25} & Open, non-neural baseline & N/A & N/A &  \\
\texttt{text-embedding-3-small} & Closed & Undisclosed & 1536 & Single \\
\texttt{text-embedding-3-large} & Closed & Undisclosed & 3072 & Single \\
\texttt{Qwen3-Embedding-0.6B} & Open & 0.6B & 1024 & Single \\
\texttt{Qwen3-Embedding-4B} & Open & 4B & 2560 & Single \\
\texttt{jina-colbert-v2} & Open & 0.56B & 128 & Multi \\
\bottomrule
\end{tabular}
\end{adjustbox}
\caption{Retrievers used in our experiments.}
\label{tab:appendix_retrievers}
\end{table}

\begin{table}[h]
\centering
\begin{adjustbox}{max width=\linewidth}
\begin{tabular}{l l l}
\toprule
Language Model & Closed/Open & Param Size \\
\midrule
\texttt{Qwen/Qwen3-4B-Instruct-2507} & Open & 4B \\
\texttt{Qwen/Qwen3-VL-2B-Instruct} & Open & 2B \\
GPT-5.4-mini & Closed & Undisclosed \\
Claude Haiku 4.5 & Closed & Undisclosed \\
GPT-5.4 & Closed & Undisclosed \\
Claude Sonnet 4.6 & Closed & Undisclosed \\
\bottomrule
\end{tabular}
\end{adjustbox}
\caption{Language models used in our experiments. Parameter counts for closed-weight models are not publicly disclosed.}
\label{tab:appendix_lms}
\end{table}

\subsection{Evaluation}
We evaluate on both code and visual document retrieval datasets. We assume an in-domain supervised setting. We
randomly sample 20\% of the benchmark queries for finetuning/training and maintain consistent splits across all experiments. Importantly, this split provides limited direct supervision relative to what is evaluated. Only 21\% of overall evaluated documents are relevant to any training query, and only 4 \% of held-out queries have a relevant document that also received supervision during training. Because the policy is applied to the entire collection after training, the large majority of evaluated documents were rewritten by a policy that never received reward signal for them. Test performance therefore primarily reflects the policy's ability to generalize a learned rewriting strategy to unsupervised documents and held-out queries, rather than memorization of training targets. Following standard practice in retriever training, for each document we use its associated positive queries as $Q_i^{+}$ and construct $Q_i^{-}$ by mining 5 hard negative queries from the positive queries associated with other documents in the training split. We use this setting throughout the main experiments. We train a separate policy for each dataset and retriever, initialized from the same instruction-tuned checkpoint. We evaluate the final policy checkpoint after 150 optimization steps. We use a fixed set of policy-training hyperparameters across datasets and retrievers, without test queries leakage. The code datasets (Table~\ref{app:datasets_code}) are drawn from MTEB and provide diverse programming tasks with natural language queries. The visual datasets (Table~\ref{app:datasets_VDR}), from Vidore2, consist of document collections across multiple domains, enabling evaluation in multimodal retrieval settings.
\label{app:datasets}
\begin{table}[H]
\centering
\begin{tabular}{l r r}
\toprule
Dataset & Num Docs & Num Queries \\
\midrule
\texttt{HumanEvalRetrieval} & 158 & 127 test / 31 train \\
\texttt{MBPPRetrieval} & 974 & 779 test / 195 train \\
\texttt{DS1000Retrieval} & 1998 & 1591 test / 407 train \\
\bottomrule
\end{tabular}
\caption{Code datasets (MTEB) used in our experiments.}
\label{app:datasets_code}
\end{table}

\begin{table}[H]
\centering
\begin{tabular}{l r r}
\toprule
Dataset & Num Docs & Num Queries \\
\midrule
\texttt{biomedical\_lectures\_v2} & 1016 & 513 test / 128 train \\
\texttt{economics\_reports\_v2} & 452 & 186 test / 46 train \\
\texttt{esg\_reports\_v2} & 1538 & 183 test / 45 train \\
\texttt{esg\_reports\_human\_labeled\_v2} & 1538 & 42 test / 10 train \\
\bottomrule
\end{tabular}
\caption{Visual datasets (Vidore2) used in our experiments.}
\label{app:datasets_VDR}
\end{table}

\begin{table}[H]
\centering
\begin{tabular}{l r r}
\toprule
Dataset & Num Docs & Num Queries \\
\midrule
\texttt{biology} & 57,359 & 83 test / 20 train \\
\texttt{earth\_science} & 121,249 & 93 test / 23 train \\
\texttt{economics} & 50,220 & 83 test / 20 train \\
\texttt{psychology} & 52,835 & 81 test / 20 train \\
\texttt{robotics} & 61,961 & 81 test / 20 train \\
\texttt{sustainable\_living} & 60,792 & 87 test / 21 train \\
\texttt{stackoverflow} & 107,081 & 94 test / 23 train \\
\bottomrule
\end{tabular}
\caption{StackExchange datasets from BRIGHT used in our experiments.}
\label{app:datasets_bright}
\end{table}

\subsubsection{Hyperparameters: Retrieval, and vLLM}
\label{final_Hyperparameters}
Unless otherwise stated, we use the same hyperparameters across MTEB and Vidore2 runs.

\paragraph{vLLM Generation}
\begin{itemize}
\item \texttt{vllm\_batch\_size} = 128
\item \texttt{max\_queries\_tokens} = 8192
\item \texttt{max\_document\_tokens} = 4096
\item \texttt{temperature} = 0.3
\item \texttt{top\_p} = 0.9
\item \texttt{top\_k} = 30
\item \texttt{repetition\_penalty} = 1.1
\item \texttt{presence\_penalty} = 0.0
\end{itemize}

\paragraph{Retrieval and Rewards}
\begin{itemize}
\item ColBERT settings: \texttt{colbert\_doc\_maxlen} = 524, \texttt{colbert\_query\_maxlen} = 64
\item \texttt{retriever\_api\_batch\_size} = 64
\item \texttt{retriever\_batch\_size} = 32

\end{itemize}

\section{Synthetic Queries}
\label{app:synthetic}

We use the frozen Qwen3-VL-2B-Instruct model to generate English
search queries from page images, without exposing it to benchmark queries or relevance labels. We restrict generation to pages associated with the labeled training split and subsample the synthetic queries to match its query budget and approximate its per-page distribution. Each source page is treated as relevant, and hard-negative queries
are mined exclusively from the synthetic bank. We retain the same retrieval corpus, training configuration, and 150-step optimization budget, and evaluate the final policy on the same held-out human
queries.

\section{Additional Plots}
\label{app:additional plots}

\begin{figure}[t]
    \centering
    \begin{adjustbox}{width=0.85\linewidth,center}
    \begin{minipage}{\linewidth}
    
    \begin{subfigure}{\linewidth}
        \centering
        \includegraphics[width=\linewidth]{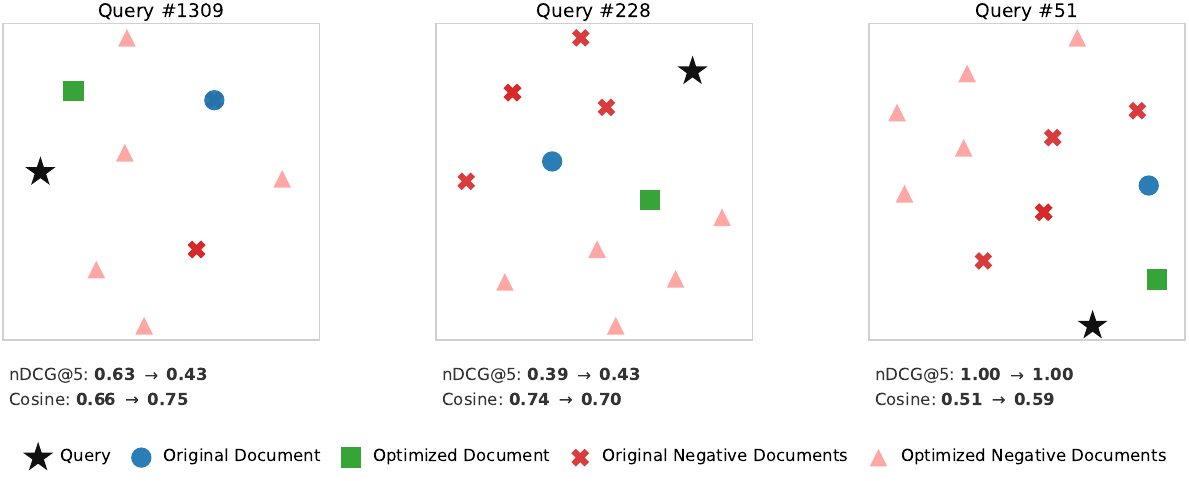}
    \end{subfigure}
    
    \vspace{0.5em}
    
    \begin{subfigure}{\linewidth}
        \centering
        \includegraphics[width=\linewidth]{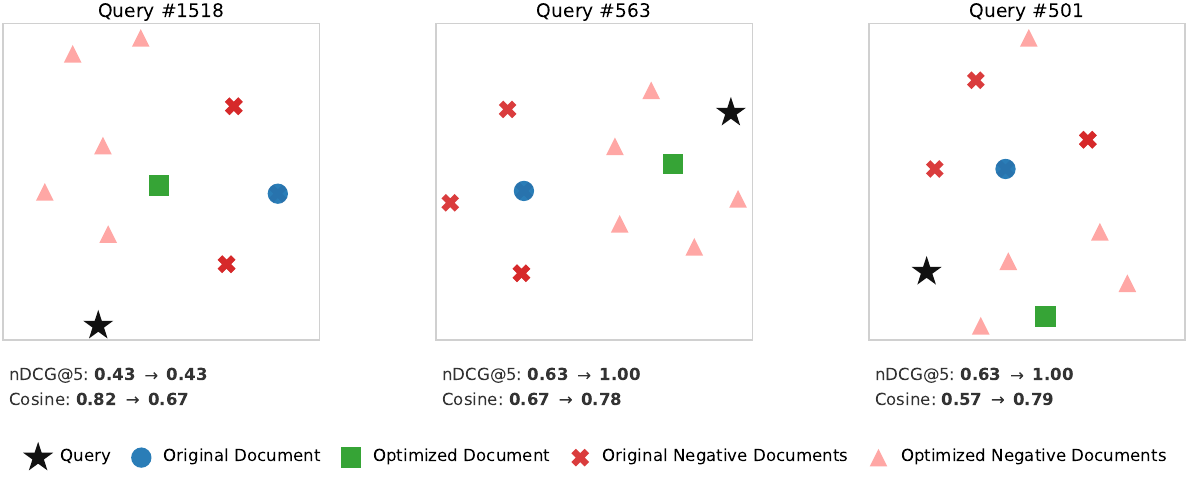}
    \end{subfigure}
    
    \vspace{0.5em}
    
    \begin{subfigure}{\linewidth}
        \centering
        \includegraphics[width=\linewidth]{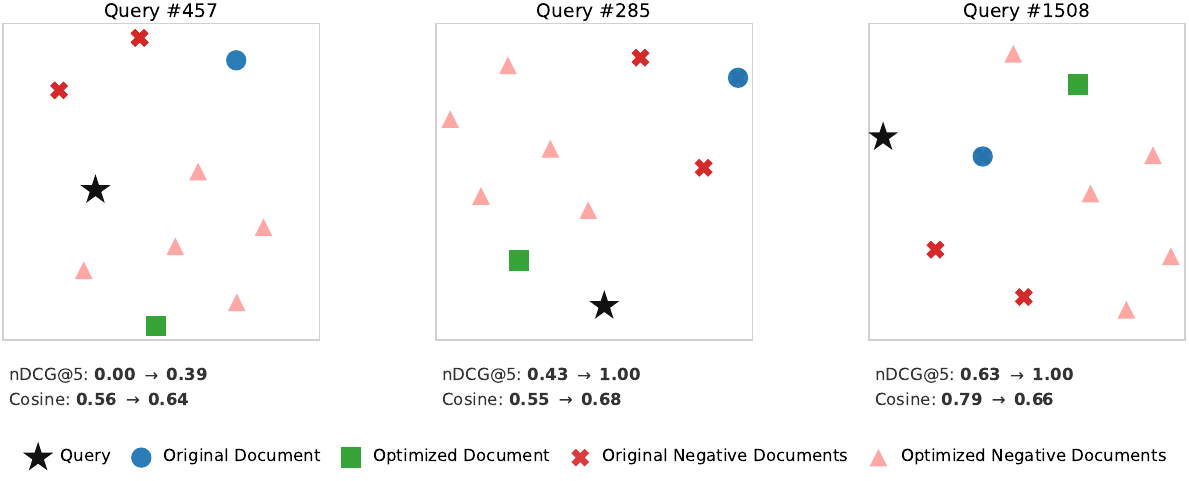}
    \end{subfigure}
    
    \vspace{0.5em}
    
    \begin{subfigure}{\linewidth}
        \centering
        \includegraphics[width=\linewidth]{images/embeddings_plots/group_007.pdf}
    \end{subfigure}
    
    \end{minipage}
    \end{adjustbox}
    \caption{Two-dimensional t-SNE projections of Qwen3-Embedding-4B embeddings across representative queries. Each plot shows a single query with its original and optimized documents, along with corresponding negatives.}
    \label{fig:embs}
\end{figure}

\clearpage
\section{Per-Domain Results}
\label{app:per-domain-res}
\begin{table}[t]
\centering
\small
\setlength{\tabcolsep}{3pt}
\renewcommand{\arraystretch}{1.08}
\begin{adjustbox}{width=\columnwidth}
\begin{tabular}{
ll
S[table-format=2.1,table-column-width=9mm] >{\centering\arraybackslash}p{7mm}
S[table-format=2.1,table-column-width=9mm] >{\centering\arraybackslash}p{7mm}
S[table-format=2.1,table-column-width=9mm] >{\centering\arraybackslash}p{7mm}
S[table-format=2.2,table-column-width=10mm] >{\centering\arraybackslash}p{10mm}
}
\toprule
\textbf{Retriever} & \textbf{Method}
& \multicolumn{2}{c}{\makecell[c]{\textbf{Human}\\\textbf{Eval}}}
& \multicolumn{2}{c}{\textbf{MBPP}}
& \multicolumn{2}{c}{\textbf{DS10K}}
& \multicolumn{2}{c}{\textbf{AVG}} \\
\cmidrule(lr){3-4}\cmidrule(lr){5-6}\cmidrule(lr){7-8}\cmidrule(lr){9-10}
 &  & \textbf{val} & {\scriptsize$\Delta$} & \textbf{val} & {\scriptsize$\Delta$} & \textbf{val} & {\scriptsize$\Delta$} & \textbf{val} & {\scriptsize$\Delta$} \\
\midrule
\multicolumn{10}{@{}l}{\textsc{Lexical}} \\
\cmidrule(lr){1-10}
\multirow{5}{*}{\hspace{0.5em}BM25} & \cellcolor{retrieverbg!85} Direct Retrieval
& 20.4 & \deltacell{\zerodelta}
& 2.4 & \deltacell{\zerodelta}
& 22.9 & \deltacell{\zerodelta}
& 15.23 & \deltacell{\zerodelta} \\
 & \cellcolor{basebg!85} GPT-5.4-mini
& 51.4 & \deltacell{\pos{+31.0}}
& 33.4 & \deltacell{\pos{+31.0}}
& 28.8 & \deltacell{\pos{+5.9}}
& 37.90 & \deltacell{\pos{+22.67}} \\
 & \cellcolor{basebg!85} Claude Haiku 4.5
& 64.7 & \deltacell{\pos{+44.3}}
& 42.0 & \deltacell{\pos{+39.6}}
& 32.4 & \deltacell{\pos{+9.5}}
& 46.37 & \deltacell{\pos{+31.13}} \\
 & \cellcolor{basebg!85} Qwen3-4B-Instruct-2507
& 72.1 & \deltacell{\pos{+51.7}}
& 51.2 & \deltacell{\pos{+48.8}}
& 35.8 & \deltacell{\pos{+12.9}}
& 53.03 & \deltacell{\pos{+37.80}} \\
 & \cellcolor{optbg!85} DocOpt
& \multicolumn{1}{c}{\textbf{80.5}} & \deltacell{\pos{+60.1}}
& \multicolumn{1}{c}{\textbf{57.7}} & \deltacell{\pos{+55.3}}
& \multicolumn{1}{c}{\textbf{36.4}} & \deltacell{\pos{+13.5}}
& \multicolumn{1}{c}{\textbf{58.20}} & \deltacell{\pos{+42.97}} \\
\midrule
\multicolumn{10}{@{}l}{\textsc{Closed-weights}} \\
\cmidrule(lr){1-10}
\multirow{5}{*}{\hspace{0.5em}OpenAI text-emb-small} & \cellcolor{retrieverbg!85} Direct Retrieval
& 84.1 & \deltacell{\zerodelta}
& 74.6 & \deltacell{\zerodelta}
& 47.2 & \deltacell{\zerodelta}
& 68.63 & \deltacell{\zerodelta} \\
 & \cellcolor{basebg!85} GPT-5.4-mini
& 92.8 & \deltacell{\pos{+8.7}}
& 77.3 & \deltacell{\pos{+2.7}}
& 42.5 & \deltacell{\negd{-4.7}}
& 70.84 & \deltacell{\pos{+2.20}} \\
 & \cellcolor{basebg!85} Claude Haiku 4.5
& 91.1 & \deltacell{\pos{+7.0}}
& 79.9 & \deltacell{\pos{+5.3}}
& 44.8 & \deltacell{\negd{-2.4}}
& 71.94 & \deltacell{\pos{+3.31}} \\
 & \cellcolor{basebg!85} Qwen3-4B-Instruct-2507
& 93.6 & \deltacell{\pos{+9.5}}
& 84.0 & \deltacell{\pos{+9.4}}
& 47.8 & \deltacell{\pos{+0.6}}
& 75.13 & \deltacell{\pos{+6.50}} \\
 & \cellcolor{optbg!85} DocOpt
& \multicolumn{1}{c}{\textbf{94.6}} & \deltacell{\pos{+10.5}}
& \multicolumn{1}{c}{\textbf{85.2}} & \deltacell{\pos{+10.6}}
& \multicolumn{1}{c}{\textbf{54.9}} & \deltacell{\pos{+7.7}}
& \multicolumn{1}{c}{\textbf{78.23}} & \deltacell{\pos{+9.60}} \\
\midrule
\multirow{5}{*}{\hspace{0.5em}OpenAI text-emb-large} & \cellcolor{retrieverbg!85} Direct Retrieval
& 93.0 & \deltacell{\zerodelta}
& 84.8 & \deltacell{\zerodelta}
& 55.4 & \deltacell{\zerodelta}
& 77.73 & \deltacell{\zerodelta} \\
 & \cellcolor{basebg!85} GPT-5.4-mini
& 96.4 & \deltacell{\pos{+3.4}}
& 83.9 & \deltacell{\negd{-0.9}}
& 47.1 & \deltacell{\negd{-8.3}}
& 75.79 & \deltacell{\negd{-1.94}} \\
 & \cellcolor{basebg!85} Claude Haiku 4.5
& \multicolumn{1}{c}{\textbf{96.9}} & \deltacell{\pos{+3.9}}
& 84.6 & \deltacell{\negd{-0.2}}
& 50.1 & \deltacell{\negd{-5.3}}
& 77.22 & \deltacell{\negd{-0.52}} \\
 & \cellcolor{basebg!85} Qwen3-4B-Instruct-2507
& 94.8 & \deltacell{\pos{+1.8}}
& \multicolumn{1}{c}{\textbf{87.2}} & \deltacell{\pos{+2.4}}
& 54.1 & \deltacell{\negd{-1.3}}
& 78.70 & \deltacell{\pos{+0.97}} \\
 & \cellcolor{optbg!85} DocOpt
& 95.1 & \deltacell{\pos{+2.1}}
& \multicolumn{1}{c}{\textbf{87.2}} & \deltacell{\pos{+2.4}}
& \multicolumn{1}{c}{\textbf{63.3}} & \deltacell{\pos{+7.9}}
& \multicolumn{1}{c}{\textbf{81.87}} & \deltacell{\pos{+4.13}} \\
\midrule
\multicolumn{10}{@{}l}{\textsc{Open-weights}} \\
\cmidrule(lr){1-10}
\multirow{7}{*}{\hspace{0.5em}Qwen3-Emb-0.6B} & \cellcolor{retrieverbg!85} Direct Retrieval
& 93.7 & \deltacell{\zerodelta}
& 87.1 & \deltacell{\zerodelta}
& 56.2 & \deltacell{\zerodelta}
& 79.00 & \deltacell{\zerodelta} \\
 & \cellcolor{basebg!85} GPT-5.4-mini
& \multicolumn{1}{c}{\textbf{98.4}} & \deltacell{\pos{+4.7}}
& 88.6 & \deltacell{\pos{+1.5}}
& 52.5 & \deltacell{\negd{-3.7}}
& 79.83 & \deltacell{\pos{+0.83}} \\
 & \cellcolor{basebg!85} Claude Haiku 4.5
& 97.7 & \deltacell{\pos{+4.0}}
& 89.7 & \deltacell{\pos{+2.6}}
& 52.8 & \deltacell{\negd{-3.4}}
& 80.07 & \deltacell{\pos{+1.07}} \\
 & \cellcolor{basebg!85} Qwen3-4B-Instruct-2507
& 96.3 & \deltacell{\pos{+2.6}}
& 89.9 & \deltacell{\pos{+2.8}}
& 55.9 & \deltacell{\negd{-0.3}}
& 80.70 & \deltacell{\pos{+1.70}} \\
 & \cellcolor{optbg!85} DocOpt
& 97.6 & \deltacell{\pos{+3.9}}
& \multicolumn{1}{c}{\textbf{90.2}} & \deltacell{\pos{+3.1}}
& 58.0 & \deltacell{\pos{+1.8}}
& 81.93 & \deltacell{\pos{+2.93}} \\
 & \cellcolor{ftbg!85} Retriever Fine-tuning
& 96.1 & \deltacell{\pos{+2.4}}
& 87.6 & \deltacell{\pos{+0.5}}
& 58.9 & \deltacell{\pos{+2.7}}
& 80.87 & \deltacell{\pos{+1.87}} \\
 & \cellcolor{jointbg!85} DocOpt - Joint Adaptation
& 98.1 & \deltacell{\pos{+4.4}}
& 89.3 & \deltacell{\pos{+2.2}}
& \multicolumn{1}{c}{\textbf{60.1}} & \deltacell{\pos{+3.9}}
& \multicolumn{1}{c}{\textbf{82.50}} & \deltacell{\pos{+3.50}} \\
\midrule
\multirow{7}{*}{\hspace{0.5em}Qwen3-Emb-4B} & \cellcolor{retrieverbg!85} Direct Retrieval
& 98.1 & \deltacell{\zerodelta}
& 91.0 & \deltacell{\zerodelta}
& 61.2 & \deltacell{\zerodelta}
& 83.43 & \deltacell{\zerodelta} \\
 & \cellcolor{basebg!85} GPT-5.4-mini
& 96.1 & \deltacell{\negd{-2.0}}
& 88.1 & \deltacell{\negd{-2.9}}
& 59.7 & \deltacell{\negd{-1.5}}
& 81.31 & \deltacell{\negd{-2.12}} \\
 & \cellcolor{basebg!85} Claude Haiku 4.5
& 96.0 & \deltacell{\negd{-2.1}}
& 88.4 & \deltacell{\negd{-2.6}}
& 60.6 & \deltacell{\negd{-0.6}}
& 81.67 & \deltacell{\negd{-1.77}} \\
 & \cellcolor{basebg!85} Qwen3-4B-Instruct-2507
& 97.3 & \deltacell{\negd{-0.8}}
& 90.5 & \deltacell{\negd{-0.5}}
& 61.1 & \deltacell{\negd{-0.1}}
& 82.97 & \deltacell{\negd{-0.47}} \\
 & \cellcolor{optbg!85} DocOpt
& \multicolumn{1}{c}{\textbf{98.5}} & \deltacell{\pos{+0.4}}
& 91.6 & \deltacell{\pos{+0.6}}
& \multicolumn{1}{c}{\textbf{63.8}} & \deltacell{\pos{+2.6}}
& \multicolumn{1}{c}{\textbf{84.63}} & \deltacell{\pos{+1.20}} \\
 & \cellcolor{ftbg!85} Retriever Fine-tuning
& 98.3 & \deltacell{\pos{+0.2}}
& \multicolumn{1}{c}{\textbf{92.0}} & \deltacell{\pos{+1.0}}
& 62.2 & \deltacell{\pos{+1.0}}
& 84.17 & \deltacell{\pos{+0.73}} \\
 & \cellcolor{jointbg!85} DocOpt - Joint Adaptation
& 97.9 & \deltacell{\negd{-0.2}}
& 91.1 & \deltacell{\pos{+0.1}}
& 63.6 & \deltacell{\pos{+2.4}}
& 84.20 & \deltacell{\pos{+0.77}} \\
\midrule
\multirow{7}{*}{\hspace{0.5em}Jina-Colbert-V2} & \cellcolor{retrieverbg!85} Direct Retrieval
& 64.3 & \deltacell{\zerodelta}
& 76.2 & \deltacell{\zerodelta}
& 36.1 & \deltacell{\zerodelta}
& 58.87 & \deltacell{\zerodelta} \\
 & \cellcolor{basebg!85} GPT-5.4-mini
& 83.7 & \deltacell{\pos{+19.4}}
& 82.9 & \deltacell{\pos{+6.7}}
& 39.5 & \deltacell{\pos{+3.4}}
& 68.67 & \deltacell{\pos{+9.80}} \\
 & \cellcolor{basebg!85} Claude Haiku 4.5
& 86.0 & \deltacell{\pos{+21.7}}
& 86.3 & \deltacell{\pos{+10.1}}
& 40.8 & \deltacell{\pos{+4.7}}
& 71.02 & \deltacell{\pos{+12.15}} \\
 & \cellcolor{basebg!85} Qwen3-4B-Instruct-2507
& 87.9 & \deltacell{\pos{+23.6}}
& 85.2 & \deltacell{\pos{+9.0}}
& 38.4 & \deltacell{\pos{+2.3}}
& 70.50 & \deltacell{\pos{+11.63}} \\
 & \cellcolor{optbg!85} DocOpt
& 91.7 & \deltacell{\pos{+27.4}}
& 87.2 & \deltacell{\pos{+11.0}}
& 42.7 & \deltacell{\pos{+6.6}}
& 73.87 & \deltacell{\pos{+15.00}} \\
 & \cellcolor{ftbg!85} Retriever Fine-tuning
& 87.2 & \deltacell{\pos{+22.9}}
& 86.4 & \deltacell{\pos{+10.2}}
& 41.0 & \deltacell{\pos{+4.9}}
& 71.53 & \deltacell{\pos{+12.67}} \\
 & \cellcolor{jointbg!85} DocOpt - Joint Adaptation
& \multicolumn{1}{c}{\textbf{92.9}} & \deltacell{\pos{+28.6}}
& \multicolumn{1}{c}{\textbf{88.4}} & \deltacell{\pos{+12.2}}
& \multicolumn{1}{c}{\textbf{43.7}} & \deltacell{\pos{+7.6}}
& \multicolumn{1}{c}{\textbf{75.00}} & \deltacell{\pos{+16.13}} \\
\bottomrule
\end{tabular}
\end{adjustbox}
\caption{Code document retrieval results. Deltas are absolute changes relative to Direct Retrieval. Bold marks the best displayed value per retriever and column.}
\label{tab:code_doc_retrieval}
\end{table}
\begin{table}[t]
\centering
\small
\setlength{\tabcolsep}{3pt}
\renewcommand{\arraystretch}{1.08}
\begin{adjustbox}{width=\columnwidth}
\begin{tabular}{
ll
S[table-format=2.1,table-column-width=9mm] >{\centering\arraybackslash}p{7mm}
S[table-format=2.1,table-column-width=9mm] >{\centering\arraybackslash}p{7mm}
S[table-format=2.1,table-column-width=9mm] >{\centering\arraybackslash}p{7mm}
S[table-format=2.1,table-column-width=9mm] >{\centering\arraybackslash}p{7mm}
S[table-format=2.2,table-column-width=10mm] >{\centering\arraybackslash}p{10mm}
}
\toprule
\textbf{Retriever} & \textbf{Method}
& \multicolumn{2}{c}{\textbf{Biomed}}
& \multicolumn{2}{c}{\textbf{Econ}}
& \multicolumn{2}{c}{\makecell[c]{\textbf{ESG}\\\textbf{Human}}}
& \multicolumn{2}{c}{\makecell[c]{\textbf{ESG}\\\textbf{Full}}}
& \multicolumn{2}{c}{\textbf{AVG}} \\
\cmidrule(lr){3-4}\cmidrule(lr){5-6}\cmidrule(lr){7-8}\cmidrule(lr){9-10}\cmidrule(lr){11-12}
 &  & \textbf{val} & {\scriptsize$\Delta$} & \textbf{val} & {\scriptsize$\Delta$} & \textbf{val} & {\scriptsize$\Delta$} & \textbf{val} & {\scriptsize$\Delta$} & \textbf{val} & {\scriptsize$\Delta$} \\
\midrule
\multicolumn{12}{@{}l}{\textsc{Lexical}} \\
\cmidrule(lr){1-12}
\multirow{5}{*}{\hspace{0.5em}BM25} & \cellcolor{retrieverbg!85} Direct Retrieval
& 12.3 & \deltacell{\zerodelta}
& 13.7 & \deltacell{\zerodelta}
& 19.3 & \deltacell{\zerodelta}
& 11.2 & \deltacell{\zerodelta}
& 14.13 & \deltacell{\zerodelta} \\
 & \cellcolor{basebg!85} GPT-5.4-mini
& 15.3 & \deltacell{\pos{+3.0}}
& 15.6 & \deltacell{\pos{+1.9}}
& 15.2 & \deltacell{\negd{-4.1}}
& \multicolumn{1}{c}{\textbf{17.5}} & \deltacell{\pos{+6.3}}
& 15.91 & \deltacell{\pos{+1.78}} \\
 & \cellcolor{basebg!85} Claude Haiku 4.5
& 17.1 & \deltacell{\pos{+4.8}}
& \multicolumn{1}{c}{\textbf{16.2}} & \deltacell{\pos{+2.5}}
& 23.0 & \deltacell{\pos{+3.7}}
& 16.7 & \deltacell{\pos{+5.5}}
& 18.25 & \deltacell{\pos{+4.13}} \\
 & \cellcolor{basebg!85} Qwen3-VL-2B-Instruct
& 16.7 & \deltacell{\pos{+4.4}}
& 15.3 & \deltacell{\pos{+1.6}}
& 28.1 & \deltacell{\pos{+8.8}}
& 12.1 & \deltacell{\pos{+0.9}}
& 18.05 & \deltacell{\pos{+3.93}} \\
 & \cellcolor{optbg!85} DocOpt
& \multicolumn{1}{c}{\textbf{17.2}} & \deltacell{\pos{+4.9}}
& 16.1 & \deltacell{\pos{+2.4}}
& \multicolumn{1}{c}{\textbf{31.5}} & \deltacell{\pos{+12.2}}
& 14.9 & \deltacell{\pos{+3.7}}
& \multicolumn{1}{c}{\textbf{19.93}} & \deltacell{\pos{+5.80}} \\
\midrule
\multicolumn{12}{@{}l}{\textsc{Closed-weights}} \\
\cmidrule(lr){1-12}
\multirow{5}{*}{\hspace{0.5em}OpenAI text-emb-small} & \cellcolor{retrieverbg!85} Direct Retrieval
& 48.8 & \deltacell{\zerodelta}
& 47.2 & \deltacell{\zerodelta}
& 47.4 & \deltacell{\zerodelta}
& 48.9 & \deltacell{\zerodelta}
& 48.08 & \deltacell{\zerodelta} \\
 & \cellcolor{basebg!85} GPT-5.4-mini
& 54.3 & \deltacell{\pos{+5.5}}
& 46.4 & \deltacell{\negd{-0.8}}
& 46.6 & \deltacell{\negd{-0.8}}
& 56.0 & \deltacell{\pos{+7.1}}
& 50.83 & \deltacell{\pos{+2.75}} \\
 & \cellcolor{basebg!85} Claude Haiku 4.5
& \multicolumn{1}{c}{\textbf{56.3}} & \deltacell{\pos{+7.5}}
& 48.3 & \deltacell{\pos{+1.1}}
& 59.2 & \deltacell{\pos{+11.8}}
& 58.5 & \deltacell{\pos{+9.6}}
& 55.57 & \deltacell{\pos{+7.49}} \\
 & \cellcolor{basebg!85} Qwen3-VL-2B-Instruct
& 53.8 & \deltacell{\pos{+5.0}}
& 47.0 & \deltacell{\negd{-0.2}}
& 55.6 & \deltacell{\pos{+8.2}}
& 56.9 & \deltacell{\pos{+8.0}}
& 53.33 & \deltacell{\pos{+5.25}} \\
 & \cellcolor{optbg!85} DocOpt
& 55.5 & \deltacell{\pos{+6.7}}
& \multicolumn{1}{c}{\textbf{51.4}} & \deltacell{\pos{+4.2}}
& \multicolumn{1}{c}{\textbf{61.9}} & \deltacell{\pos{+14.5}}
& \multicolumn{1}{c}{\textbf{61.8}} & \deltacell{\pos{+12.9}}
& \multicolumn{1}{c}{\textbf{57.65}} & \deltacell{\pos{+9.58}} \\
\midrule
\multirow{5}{*}{\hspace{0.5em}OpenAI text-emb-large} & \cellcolor{retrieverbg!85} Direct Retrieval
& 54.3 & \deltacell{\zerodelta}
& 52.8 & \deltacell{\zerodelta}
& 56.8 & \deltacell{\zerodelta}
& 54.4 & \deltacell{\zerodelta}
& 54.58 & \deltacell{\zerodelta} \\
 & \cellcolor{basebg!85} GPT-5.4-mini
& \multicolumn{1}{c}{\textbf{60.3}} & \deltacell{\pos{+6.0}}
& 48.5 & \deltacell{\negd{-4.3}}
& 57.1 & \deltacell{\pos{+0.3}}
& 57.3 & \deltacell{\pos{+2.9}}
& 55.81 & \deltacell{\pos{+1.23}} \\
 & \cellcolor{basebg!85} Claude Haiku 4.5
& 59.4 & \deltacell{\pos{+5.1}}
& 52.6 & \deltacell{\negd{-0.2}}
& 61.5 & \deltacell{\pos{+4.7}}
& 58.5 & \deltacell{\pos{+4.1}}
& 58.00 & \deltacell{\pos{+3.42}} \\
 & \cellcolor{basebg!85} Qwen3-VL-2B-Instruct
& 57.0 & \deltacell{\pos{+2.7}}
& 49.2 & \deltacell{\negd{-3.6}}
& 58.8 & \deltacell{\pos{+2.0}}
& \multicolumn{1}{c}{\textbf{63.1}} & \deltacell{\pos{+8.7}}
& 57.03 & \deltacell{\pos{+2.45}} \\
 & \cellcolor{optbg!85} DocOpt
& 60.0 & \deltacell{\pos{+5.7}}
& \multicolumn{1}{c}{\textbf{53.6}} & \deltacell{\pos{+0.8}}
& \multicolumn{1}{c}{\textbf{64.6}} & \deltacell{\pos{+7.8}}
& 59.4 & \deltacell{\pos{+5.0}}
& \multicolumn{1}{c}{\textbf{59.40}} & \deltacell{\pos{+4.83}} \\
\midrule
\multicolumn{12}{@{}l}{\textsc{Open-weights}} \\
\cmidrule(lr){1-12}
\multirow{7}{*}{\hspace{0.5em}Qwen3-Emb-0.6B} & \cellcolor{retrieverbg!85} Direct Retrieval
& 46.3 & \deltacell{\zerodelta}
& 39.8 & \deltacell{\zerodelta}
& 38.6 & \deltacell{\zerodelta}
& 33.4 & \deltacell{\zerodelta}
& 39.53 & \deltacell{\zerodelta} \\
 & \cellcolor{basebg!85} GPT-5.4-mini
& 54.5 & \deltacell{\pos{+8.2}}
& 42.6 & \deltacell{\pos{+2.8}}
& 43.5 & \deltacell{\pos{+4.9}}
& 47.0 & \deltacell{\pos{+13.6}}
& 46.92 & \deltacell{\pos{+7.39}} \\
 & \cellcolor{basebg!85} Claude Haiku 4.5
& 43.4 & \deltacell{\negd{-2.9}}
& 41.0 & \deltacell{\pos{+1.2}}
& 30.0 & \deltacell{\negd{-8.6}}
& 35.0 & \deltacell{\pos{+1.6}}
& 37.34 & \deltacell{\negd{-2.19}} \\
 & \cellcolor{basebg!85} Qwen3-VL-2B-Instruct
& 54.4 & \deltacell{\pos{+8.1}}
& 45.8 & \deltacell{\pos{+6.0}}
& 54.1 & \deltacell{\pos{+15.5}}
& 55.8 & \deltacell{\pos{+22.4}}
& 52.53 & \deltacell{\pos{+13.00}} \\
 & \cellcolor{optbg!85} DocOpt
& \multicolumn{1}{c}{\textbf{58.5}} & \deltacell{\pos{+12.2}}
& \multicolumn{1}{c}{\textbf{53.9}} & \deltacell{\pos{+14.1}}
& 58.2 & \deltacell{\pos{+19.6}}
& 58.8 & \deltacell{\pos{+25.4}}
& \multicolumn{1}{c}{\textbf{57.35}} & \deltacell{\pos{+17.83}} \\
 & \cellcolor{ftbg!85} Retriever Fine-tuning
& 53.0 & \deltacell{\pos{+6.7}}
& 47.8 & \deltacell{\pos{+8.0}}
& \multicolumn{1}{c}{\textbf{65.0}} & \deltacell{\pos{+26.4}}
& 59.3 & \deltacell{\pos{+25.9}}
& 56.28 & \deltacell{\pos{+16.75}} \\
 & \cellcolor{jointbg!85} DocOpt - Joint Adaptation
& 55.2 & \deltacell{\pos{+8.9}}
& 49.5 & \deltacell{\pos{+9.7}}
& 52.5 & \deltacell{\pos{+13.9}}
& \multicolumn{1}{c}{\textbf{60.1}} & \deltacell{\pos{+26.7}}
& 54.33 & \deltacell{\pos{+14.80}} \\
\midrule
\multirow{7}{*}{\hspace{0.5em}Qwen3-Emb-4B} & \cellcolor{retrieverbg!85} Direct Retrieval
& 51.2 & \deltacell{\zerodelta}
& 45.3 & \deltacell{\zerodelta}
& 42.7 & \deltacell{\zerodelta}
& 38.8 & \deltacell{\zerodelta}
& 44.50 & \deltacell{\zerodelta} \\
 & \cellcolor{basebg!85} GPT-5.4-mini
& 50.9 & \deltacell{\negd{-0.3}}
& 46.7 & \deltacell{\pos{+1.4}}
& 41.1 & \deltacell{\negd{-1.6}}
& 52.7 & \deltacell{\pos{+13.9}}
& 47.83 & \deltacell{\pos{+3.33}} \\
 & \cellcolor{basebg!85} Claude Haiku 4.5
& 46.2 & \deltacell{\negd{-5.0}}
& 45.4 & \deltacell{\pos{+0.1}}
& 37.9 & \deltacell{\negd{-4.8}}
& 44.1 & \deltacell{\pos{+5.3}}
& 43.39 & \deltacell{\negd{-1.11}} \\
 & \cellcolor{basebg!85} Qwen3-VL-2B-Instruct
& 58.8 & \deltacell{\pos{+7.6}}
& 51.5 & \deltacell{\pos{+6.2}}
& 57.1 & \deltacell{\pos{+14.4}}
& 57.6 & \deltacell{\pos{+18.8}}
& 56.25 & \deltacell{\pos{+11.75}} \\
 & \cellcolor{optbg!85} DocOpt
& 58.9 & \deltacell{\pos{+7.7}}
& 53.9 & \deltacell{\pos{+8.6}}
& 60.0 & \deltacell{\pos{+17.3}}
& 58.0 & \deltacell{\pos{+19.2}}
& 57.70 & \deltacell{\pos{+13.20}} \\
 & \cellcolor{ftbg!85} Retriever Fine-tuning
& 58.9 & \deltacell{\pos{+7.7}}
& \multicolumn{1}{c}{\textbf{55.0}} & \deltacell{\pos{+9.7}}
& \multicolumn{1}{c}{\textbf{60.8}} & \deltacell{\pos{+18.1}}
& \multicolumn{1}{c}{\textbf{62.2}} & \deltacell{\pos{+23.4}}
& \multicolumn{1}{c}{\textbf{59.23}} & \deltacell{\pos{+14.73}} \\
 & \cellcolor{jointbg!85} DocOpt - Joint Adaptation
& \multicolumn{1}{c}{\textbf{59.6}} & \deltacell{\pos{+8.4}}
& 51.5 & \deltacell{\pos{+6.2}}
& 58.6 & \deltacell{\pos{+15.9}}
& \multicolumn{1}{c}{\textbf{62.2}} & \deltacell{\pos{+23.4}}
& 57.98 & \deltacell{\pos{+13.48}} \\
\midrule
\multirow{7}{*}{\hspace{0.5em}Jina-Colbert-V2} & \cellcolor{retrieverbg!85} Direct Retrieval
& 55.9 & \deltacell{\zerodelta}
& 51.6 & \deltacell{\zerodelta}
& 51.5 & \deltacell{\zerodelta}
& 56.0 & \deltacell{\zerodelta}
& 53.75 & \deltacell{\zerodelta} \\
 & \cellcolor{basebg!85} GPT-5.4-mini
& 59.5 & \deltacell{\pos{+3.5}}
& 48.9 & \deltacell{\negd{-2.7}}
& 50.4 & \deltacell{\negd{-1.1}}
& 52.5 & \deltacell{\negd{-3.5}}
& 52.81 & \deltacell{\negd{-0.94}} \\
 & \cellcolor{basebg!85} Claude Haiku 4.5
& 59.9 & \deltacell{\pos{+4.0}}
& 49.1 & \deltacell{\negd{-2.5}}
& 46.9 & \deltacell{\negd{-4.6}}
& 56.3 & \deltacell{\pos{+0.3}}
& 53.05 & \deltacell{\negd{-0.71}} \\
 & \cellcolor{basebg!85} Qwen3-VL-2B-Instruct
& 57.8 & \deltacell{\pos{+1.9}}
& 52.4 & \deltacell{\pos{+0.8}}
& 56.7 & \deltacell{\pos{+5.2}}
& 56.4 & \deltacell{\pos{+0.4}}
& 55.83 & \deltacell{\pos{+2.07}} \\
 & \cellcolor{optbg!85} DocOpt
& 58.9 & \deltacell{\pos{+3.0}}
& 54.2 & \deltacell{\pos{+2.6}}
& 63.1 & \deltacell{\pos{+11.6}}
& 55.9 & \deltacell{\negd{-0.1}}
& 58.03 & \deltacell{\pos{+4.27}} \\
 & \cellcolor{ftbg!85} Retriever Fine-tuning
& 59.3 & \deltacell{\pos{+3.4}}
& 53.2 & \deltacell{\pos{+1.6}}
& 59.4 & \deltacell{\pos{+7.9}}
& 55.1 & \deltacell{\negd{-0.9}}
& 56.75 & \deltacell{\pos{+3.00}} \\
 & \cellcolor{jointbg!85} DocOpt - Joint Adaptation
& \multicolumn{1}{c}{\textbf{61.2}} & \deltacell{\pos{+5.3}}
& \multicolumn{1}{c}{\textbf{61.0}} & \deltacell{\pos{+9.4}}
& \multicolumn{1}{c}{\textbf{72.6}} & \deltacell{\pos{+21.1}}
& \multicolumn{1}{c}{\textbf{58.4}} & \deltacell{\pos{+2.4}}
& \multicolumn{1}{c}{\textbf{63.30}} & \deltacell{\pos{+9.55}} \\
\bottomrule
\end{tabular}
\end{adjustbox}
\caption{Visual document retrieval results. Direct Retrieval uses OCR. Deltas are absolute changes relative to Direct Retrieval.}
\label{tab:visual_doc_retrieval}
\end{table}

\end{document}